\documentclass[10pt]{article}

\usepackage{etoolbox}
\newtoggle{frontierspaper}

\togglefalse{frontierspaper}

\usepackage{amsmath, amssymb, epsfig}
\usepackage{mathrsfs}
\usepackage{tikz}
\usepackage{algorithm}
\usepackage{algpseudocode}
\usepackage{bm}

\iftoggle{frontierspaper}{
\usepackage{float}
\newfloat{algorithm}{t}{lop}
\usepackage{url,lineno,microtype}
\usepackage[onehalfspacing]{setspace}
\usepackage[nomarkers]{endfloat}

\linenumbers
}{\usepackage{natbib}
\usepackage[preprint]{tmlr}
}

\algnewcommand\algorithmicinput{\textbf{Input}}
\algnewcommand\algorithmicoutput{\textbf{Output}}
\algnewcommand\INPUT{\item[\algorithmicinput]}
\algnewcommand\OUTPUT{\item[\algorithmicoutput]}

\usepackage[hidelinks]{hyperref}
\usepackage[capitalize]{cleveref}

\newtheorem{theorem}{Theorem}[section]

\newtheorem{definition}[theorem]{Definition}

\newtheorem{remark}[theorem]{Remark}

\DeclareMathOperator*{\argmin}{arg min} 
\DeclareMathOperator*{\argmax}{arg max} 
\DeclareMathOperator*{\rank}{rank}

\usepackage{xspace}
\newcommand{\fnmf}{Fairer-NMF\xspace}

\newcommand{\mat}[1]{\bm{#1}}
\newcommand{\R}{\mathbb{R}}

\iftoggle{frontierspaper}{
\def\keyFont{\fontsize{8}{11}\helveticabold }
\def\firstAuthorLast{Kassab {et~al.}}
\def\Authors{Lara Kassab\,$^{1}$, Erin George\,$^{2,*}$, Deanna Needell\,$^{3}$, Haowen Geng\,$^{4}$, Nika Jafar Nia\,$^{5}$, and Aoxi Li\,$^{6}$}

}

{

\title{Towards a Fairer Non-negative Matrix Factorization}
\author{\name Lara Kassab \email lkassab@fullerton.edu \\
\addr Department of Mathematics, California State University, Fullerton
\AND
\name Erin George \email e2george@ucsd.edu \\
\addr Department of Mathematics, University of California, San Diego
\AND
\name Deanna Needell \email deanna@math.ucla.edu\\
\addr Department of Mathematics, University of California, Los Angeles
\AND
\name Haowen Geng \email haowengeng2025@u.northwestern.edu\\
\addr McCormick School of Engineering \\ Northwestern University
\AND
\name Nika Jafar Nia \email njafarnia24@amherst.edu\\
\addr Department of Mathematics \\ Amherst College
\AND
\name Aoxi Li \email aoxi@berkeley.edu \\
\addr Department of Industrial Engineering and Operations Research \\ Univ. of California, Berkeley
}
    
}    
\date{}

\begin{document}
\iftoggle{frontierspaper}{
\onecolumn
\firstpage{1}

\title{Fairer Non-negative Matrix Factorization}
\author[\firstAuthorLast ]{\Authors}
\address{}
\correspondance{}
\extraAuth{}
}{}
\maketitle
\begin{abstract}
There has been a recent critical need to study fairness and bias in machine learning (ML) algorithms. Since there is clearly no one-size-fits-all solution to fairness, ML methods should be developed alongside bias mitigation strategies that are practical and approachable to the practitioner. Motivated by recent work on ``fair" PCA, here we consider the more challenging method of non-negative matrix factorization (NMF) as both a showcasing example and a method that is important in its own right for both topic modeling tasks and feature extraction for other ML tasks. 
We demonstrate that a modification of the objective function, by using a min-max formulation, may \textit{sometimes} be able to offer an improvement in fairness for groups in the population.  
We derive two methods for the objective minimization, a multiplicative update rule as well as an alternating minimization scheme, and discuss implementation practicalities. We include a suite of synthetic and real experiments that show how the method may improve fairness while also highlighting the important fact that this may sometime increase error for some individuals and fairness is not a rigid definition and method choice should strongly depend on the application at hand.
\iftoggle{frontierspaper}{

\tiny
 \keyFont{ \section{Keywords:} fairness, non-negative matrix factorization, dimensionality reduction, Fairer-NMF, topic modeling}}{}
\end{abstract}

\section{Introduction}
\label{sec:Introduction} 
Machine Learning (ML) and Artificial Intelligence (AI) have seen a huge surge in uses and applications in recent times and are being used in nearly every aspect of society. Despite this, there are serious and critical issues that propagate bias, disseminate unfair outcomes, and affect racial and social justice~\citep{chatgptbias,verge1,verge2}.   
This lack of equity typically stems from many sources, from bias in the data to algorithmic bias and even post-processing decisions~\citep{barocas2023fairness,mehrabi2021survey}.  
In this work, we focus on inequities typically stemming from both the lack of fair representation of the data as well as algorithmic bias in the treatment of that data. 
 ML techniques are used in a wide array of  applications ranging from medical diagnostics and predictions to recidivism and sentencing decisions.  
At the heart of these applications lies the need for an algorithm to uncover hidden themes or features, that either explain some studied medical phenomenon or are used downstream for a learning task like classification or prediction.   
Because the data itself is often biased and ML methods tend to be designed to perform well \textit{on average}, this can lead to misclassification and poorly explained patterns, particularly for underrepresented populations.

In this work, we focus on addressing this problem for topic modeling, or more generally, (unsupervised) dimensionality reduction tasks. 
We specifically consider non-negative matrix factorization (NMF), a powerful method used in a wide array of ML applications.
It serves as a mathematically tangible example of a method for discovering data trends that highlight the inequities we wish to study~\citep{lee2001algorithms,lee1999learning}. 
Indeed, at the core of NMF is a simple objective function that asks that the factorization has a small \textit{average reconstruction error}. Because one only minimizes such an average, it is clear that population subgroups of small size may often be overshadowed (i.e. experience large reconstruction error) even when the total error is quite small relative to the population size. If NMF is being used to study a population's features and the desire is that \textit{all} members or groups of the populations be represented in those features, this is clearly detrimental.  Further, if NMF is being used for feature identification along with e.g. classification or some other ML task, this example would imply that those small subgroups experience incredible inaccuracy while the majority of the population benefits from accurate predictions.  Especially when such a method is used without a fairness analysis, these may go unnoticed and cause extreme harm, such as is the case in medical, criminal justice, and many other applications.

This work aims to explore a fairer alternative objective function for NMF under a specific framework of fairness and present algorithmic implementations for solving this formulation.
Before one can seek to promote fairness, it is essential to first define what fairness means.
Defining fairness and fighting against bias and discrimination have existed long before the advent of machine learning~\citep{saxena2019perceptions}. 
Notably, definitions of fairness vary across tasks and settings and are generally non-universal which contributes to the difficulty of solving such problems~\citep{barocas2023fairness,mehrabi2021survey}.
We emphasize here, as we do in the title, that our goal must be humble; we seek a \textit{fairer---not fair---}formulation, and even that will only be fairer for certain contexts and applications.  In fact, in some cases, applying a fairer method may also result in increased error (or decreased accuracy) for some individuals, meaning that the use of any method (``fairer" or not) should always be handled with care, and appropriate fairness metrics should be measured regardless.  See Section \ref{sec:discuss} for more discussion on this and other related points.
Nonetheless, we view this as an important step forward, which will hopefully lead to ML algorithms with more transparency and flexibility for the end user to identify and mitigate bias.

\section{Contributions and Organization}
\subsection{Contribution}
We view the contributions of this work in several ways. 
First, we showcase how NMF, a method used for both transparent direct data analysis and as a precursor or proxy for interpretable feature extraction for other ML methods, may often produce inequitable outcomes. 
Next, we present a min-max mitigation strategy to improve fairness that is motivated by the so-called fair variant of principal component analysis (PCA) \citep{samadi2018price}, aimed at mitigating bias stemming from group imbalance in size and complexity. 
Despite the added complications of non-negativity  in the analysis, we derive two algorithms to solve the proposed fairer formulation of NMF: a multiplicative updates scheme as well as an alternating minimization scheme. 
Lastly and perhaps most importantly, we show through a suite of synthetic and real data experiments that there are settings where this formulation improves fairness, and that also there may be situations where it may not -- depending on the desired form of fairness and application.  
We believe being able to provide this type of transparency is an absolutely critical first step towards a fairer world of ML and AI. 

\subsection{Organization}
In \cref{sec:related-works}, we present related works in fair unsupervised learning techniques, particularly dimensionality reduction.
We provide an overview of NMF, its applications, and existing algorithms in \cref{sec:background}.
In \cref{sec:fairer-NMF}, we further discuss the objective function of standard NMF at the group level and define the fairness criterion of our proposed NMF formulation called \fnmf. 
Then, in \cref{sec: fnmf-algorithms}, we present two algorithms for solving \fnmf.
Lastly, in \cref{sec:num-exp}, we present numerical experiments on synthetic and real data to demonstrate the performance of the algorithms.

\subsection{Notation}
We use boldfaced upper-case Latin letters (e.g., $\mat X$) to denote matrices and $\mat X \in \R_{\geq 0}^{m \times n}$ to denote an $m \times n$ matrix with real non-negative entries.
The Frobenius norm of a matrix $\mat X$ is denoted by $\|\mat X\|$.
The notation $\mat{A}/\mat{B}$ indicates entrywise division, $\mat A \odot \mat B$ entrywise multiplication, and $\mat A \mat B$ standard matrix multiplication. 
The notation $\bm {e}_{\ell} \in \R^L$ indicates the standard basis vector in $\R^L$ where $\ell$-th entry in the vector $\bm  {e}$ is 1 and all other entries are zero.

For a dataset partitioned into two (or more) mutually exclusive sample groups, we write the data matrix $\mat X$ with $m$ number of samples and $n$ number of features in block format as,
$$\mat X = \begin{bmatrix} \mat X _A \\ \mat X _B \end{bmatrix} \in \R_{\geq 0}^{m \times n} \quad \text{where } 
\mat X _A \in \R_{\geq 0}^{m_1 \times n}, \mat X _B \in \R_{\geq 0}^{m_2 \times n}.
$$
The matrices $\mat X _ A$ and $\mat X_ B$ are constructed from the rows in $\mat X$ corresponding to sample group $A$  and $B$, respectively.
In general, we write $| A |$ to denote the size of group $A$.
Here, $|A| = m_1$ and $|B| = m_2$ with $m_1 + m_2 = m$.
Additionally, the notation $\mat X   _A$ generally means we restrict to the rows of $\mat X$ with sample indices given by $A$.

\section{Related Works}
\label{sec:related-works}

In this section, we focus on related work that is most pertinent to our study, rather than providing an exhaustive review. We provide more technical presentation of NMF in the section following.

\subsection{Standard Non-negative Matrix Factorization}
\label{sec:nmf-intro}
\emph{Topic modeling} is a machine learning technique used to reveal latent themes or patterns from large datasets.
A popular technique for topic modeling that provides a low-rank approximation of a matrix is non-negative matrix factorization (NMF)~\citep{lee1999learning,lee2001algorithms}.
NMF has garnered increasing attention due to its effectiveness in handling large-scale data across various domains.
In image processing, NMF is employed for tasks like feature extraction and perceptual hashing~\citep{lee1999learning,rajapakse2004color,tang2013robust}. 
In the field of text mining, it has proven useful for document clustering and semantic analysis~\citep{berry2005email,xu2003document}.
In the medical field, it has been employed on applications ranging from fraud detection~\citep{zhu2011health} and phenotyping~\citep{joshi2016identifiable} to studying trends in health and disease from record or survey data~\citep{hamamoto2022application,hassaine2020learning,Johnson2024Towar,VHNL20lyme}.
Indeed, due to the non-negativity constraints, NMF acquires a parts-based, sparse representation of the data~\citep{lee1999learning}.
When the features are naturally non-negative, this approach often enhances interpretability compared to traditional methods like Principal Components Analysis (PCA)~\citep{lee1999learning}.

\subsection{Fair Unsupervised Learning}
\label{sec:related-works-unsupervised}

The topic of fairness in clustering has recently gained significant interest in the machine learning community with~\citep{chierichetti2017fair} leading the first work on fair clustering.
Due to the difficulty in defining and enforcing fairness criteria for unsupervised learning tasks, including clustering techniques, many different fairness notions for clustering exist~\citep{backurs2019scalable, chen2019proportionally,ghadiri2021socially,mahabadi2020individual}.
An overview of fair clustering is given in~\citep{chhabra2021overview}.

Fairness issues in recommender systems have also recently attracted increasing attention, leading to the emergence of works aimed at mitigating bias (e.g.,~\citep{li2021user,zhu2020measuring}).
In~\citep{wang2023survey}, the authors provide a survey on the fairness of recommender systems. 
Some works have proposed fairness-aware matrix factorizations for recommender systems (e.g.,~\citep{togashi2022fair}) including federated approaches (e.g.,~\citep{liu2022fairness}). 
We note that the works of~\citep{togashi2022fair} and~\citep{liu2022fairness} differ substantially from ours.
Both~\citep{togashi2022fair} and~\citep{liu2022fairness} are limited to recommendation and ranking contexts, focusing on fairness-aware collaborative filtering for item exposure fairness. They employ standard matrix factorization as a tool for building recommender systems.
In contrast, our work directly incorporates group fairness into the NMF formulation while serving as a general dimensionality reduction technique for diverse ML applications, including topic modeling.

While the aforementioned works focus on fairness in recommendation contexts, fairness has also been studied in broader dimensionality reduction settings. In~\citep{buet2022towards}, the authors investigate fairness of generalized low-rank models (GLRMs), including NMF, in unsupervised learning settings. 
We remark that our fairness formulation differs from~\citet{buet2022towards}, as we detail in \cref{sec:fnmf-objective}.


\subsection{Fair Principal Component Analysis}
\label{sec: fair-pca}
In~\citep{samadi2018price}, the authors investigate how PCA might inadvertently introduce bias.
The numerical experiments show that PCA incurs much higher average reconstruction error for one population than another (e.g., lower- versus higher-educated individuals), even when the populations are of similar sizes.
The authors established a formulation called, \textit{Fair PCA}, that addresses this bias under a specific framework of fairness.
For a given matrix $\mat Y \in \R^{a\times n}$ denote by $\mat {\hat Y} \in \R^{a\times n}$ the optimal rank-$d$ approximation of $\mat Y$. 
Given $\mat Z \in \R^{a\times n}$ with rank at most $d$, the reconstruction loss is defined as,
\[ loss(\mat Y, \mat Z) := \|\mat Y - \mat Z\|_F^2 - 
\|\mat Y - \mat {\hat Y}\|_F^2.\]
Consider a data matrix 
$\begin{bmatrix}
    \mat A \\ \mat B
\end{bmatrix}
\in \R^{m \times n}$
where the rows in $\mat A$ and $\mat B$ are samples in the data belonging to a group $A$ and $B$, respectively.
The problem of finding a projection into $d$-dimensions in \textit{Fair PCA} is defined as solving,
\begin{equation}
\label{eq: fair-pca}
\min _{\mat U \in \R^{m \times n}, \rank(\mat U) \leq d} \max \left \{\frac{1}{|A|}
loss(\mat A, \mat U_A),
\frac{1}{|B|} loss(\mat B, \mat U_B) \right \}.
\end{equation} 
This criterion of Fair PCA, \cref{eq: fair-pca}, seeks to minimize the maximum of the average reconstruction loss across different groups which fits under what is known as the min-max framework or \textit{social fairness}.
In~\citep{tantipongpipat2019multi}, the authors further introduce a multi-criteria dimensionality reduction problem, where multiple objectives are optimized simultaneously.
One application of this model is capturing several fairness criteria in dimensionality reduction, such as the Fair PCA problem~\citep{samadi2018price}.

Motivated by~\citep{samadi2018price}, here we ask whether a similar framework can be used for other linear algebraic-based ML approaches. 
In particular, we explore how the non-negativity constraints, which offer more interpretability than PCA but also introduce analytical challenges, can be adapted to such a framework.

\section{Standard NMF Formulation}
\label{sec:background}

Given a non-negative matrix $\mat X \in \R _{\geq 0}^{m \times n}$
and a target dimension $r \in \mathbb N$, NMF decomposes $\mat X$ into a product of two low-dimensional non-negative matrices: $\mat W  \in \R_{\geq 0}^{m \times r}$ and $\mat H   \in \R _{\geq 0}^{r \times n}$, such that
\[
\mat X \approx \mat W  \mat H  
\]
We consider $\mat X$ to be a data matrix where the rows represent data samples and the columns represent data features.
Typically, $r>0$ is chosen such that $r \ll \min\{m,n\}$ to reduce the dimension of the original data matrix or reveal hidden patterns in the data.
The matrix $\mat W$ is called the \textit{representation matrix} and $\mat H$ is called the \textit{dictionary matrix}.
The rows of $\mat H$ are generally referred to as \emph{topics}, which are characterized by features of the dataset.
Each row of $\mat W$ provides the approximate representation of the respective row in $\mat X$ in the lower-dimensional space spanned by the rows of $\mat H$.
Thus, the data points are well approximated by an additive linear combination of the topics.

We note that in the NMF literature, $r$ is referred to as the target dimension, the number of desired topics, or the desired non-negative rank. 
It is a user-specified hyperparameter and can be estimated heuristically.
The \textit{non-negative rank} of a matrix $\mat X$ is the smallest integer $r^* > 0$ such that there exists an exact NMF decomposition:
$ \mat X = \mat W  \mat H  $ where $\mat W  \in \R_{\geq 0}^{m \times r^*}$ and $\mat H   \in \R _{\geq 0}^{r^* \times n}$.
Computing the exact non-negative rank of a matrix is NP-hard~\citep{vavasis2010complexity}. 
Therefore, several formulations for the non-negative approximation, $\mat X \approx \mat W \mat H \in \R_{\geq 0}^{m \times n}$, have been studied~\citep{cichocki2009nonnegative,lee1999learning,lee2001algorithms} that seek to minimize the reconstruction error of the decomposition.

\begin{definition}[Relative Reconstruction Error] \label{def:rec_error} 
Suppose $\mat X \in \R_{\geq 0}^{m \times n}$ and $r< \min\{m,n\} \in \mathbb N$.
For a given $\mat W  \in \R_{\geq 0}^{m \times r}$ and $\mat H   \in \R _{\geq 0}^{r \times n}$ we define the \emph{reconstruction error} of $\mat X$ as
$\| \mat  X - \mat W \mat H\|$ and the \emph{relative reconstruction error} of $\mat X$ 
 as $\| \mat  X - \mat W \mat H\|/\|\mat  X\|$.
\end{definition}

One of the most popular formulations of finding an NMF approximation uses the Frobenius norm as a measure of the reconstruction error,
\begin{equation}
\label{eq:nmf_form}
\underset{\mat W \in \R_{\geq 0}^{m \times r}, \mat H \in \R_{\geq 0}^{r \times n}}{\argmin}\|\mat X - \mat W \mat H\|^2.
\end{equation}
Throughout the paper, we refer to this formulation as \textit{rank-$r$ NMF} or \textit{standard NMF with rank $r$}.
For simplicity and as common in the literature of NMF, we will refer to non-negative rank simply as rank.

Many numerical optimization techniques can be applied to find local minima for the NMF problem defined in \cref{eq:nmf_form}
(e.g.,~\citep{cichocki2009nonnegative,kim2008fast,kim2008nonnegative,lin2007projected,paatero1994positive}).  
Note that although~\cref{eq:nmf_form} is a non-convex optimization problem, it is convex in $\mat W$ when $\mat H$ is held fixed and vice-versa. 
Thus, an \textit{alternating minimization} (AM) approach (see e.g.,~\citep{bertsekas1997nonlinear}) can be used to find local minima:
\begin{align*} 
&\mat W^{(k)} \gets \underset{\mat W \in \R_{\geq 0}^{m \times r}}{\argmin}\|\mat X - \mat W \mat H^{(k)}\|^2 \\
& \mat H^{(k)} \gets \underset{\mat H \in \R_{\geq 0}^{r \times n}}{\argmin}\|\mat X - \mat W^{(k)} \mat H\|^2
\end{align*}
where $k$ denotes the $k$-th iteration.  Both of these convex problems are non-negative least squares problems, and specialized solvers exist to find solutions.

Another minimization method is the \textit{multiplicative updates} (MU) method proposed in~\citep{lee2001algorithms}.
The method can be viewed as an entrywise projected gradient descent algorithm.
The choice of stepsize for each entry of the updating matrix results in multiplicative (rather than additive) update rules that ensure non-negativity.
The algorithm performs alternating steps in updating $\mat W$ and $\mat H$:
\begin{align*} 
&\mat W^{(k)} \gets \mat W^{(k-1)} \odot \frac{\mat X {\mat H^{(k-1)}}^\top}{\mat W^{(k-1)} \mat H^{(k-1)} {\mat H^{(k-1)}}^\top} \\
& \mat H^{(k)} \gets \mat H^{(k-1)} \odot \frac{{\mat W^{(k)}}^\top \mat X}{{\mat W^{(k)}}^\top \mat W^{(k)} \mat H^{(k-1)}}
\end{align*}
The multiplicative updates algorithm is commonly used due to its ease of implementation, the absence of the need for user-defined hyperparameters, and desirable monotonicity properties~\citep{lee2001algorithms}.

We comment here that the recent work \citep{gillis2021distributionally} has developed a distributionally robust algorithm for multi-objective NMF using multiplicative updates.  
The derivation of their MU scheme follows similarly to ours (in \cref{sec:mu-scheme}), and in fact, it would be interesting future work to incorporate both our fairness objective as well as their robust framework.

\section{\fnmf Formulation}
\label{sec:fairer-NMF}
In this section, we provide a technical presentation of standard NMF, highlight characteristics of its objective function at the group level, and propose our approach, \fnmf.

\subsection{Standard NMF at the Group Level}
\label{sec: nmf-objective}
Suppose a dataset consists of two mutually exclusive groups $A$ (with size $|A| = m_1$) and $B$ (with size $|B| = m_2$).
For example, these groups could be divided based on a protected attribute in the data.
We can write the NMF of a data matrix $\mat X \in \R_{\geq 0}^{m \times n}$ as,
\begin{equation} \mat X = 
\begin{bmatrix} \mat X_A \\ \mat X_B \end{bmatrix} \approx \begin{bmatrix} \mat W_A \\ \mat W_B \end{bmatrix} \mat H \label{eq:stackedNMF}
\end{equation}
where $\mat W_A \in \R_{\geq 0}^{m_1 \times r}$ is the representation matrix corresponding to $\mat X_A$, $\mat W_B \in \R_{\geq 0}^{m_2 \times r}$ is the representation matrix corresponding to $\mat X_B$, and $\mat H \in \R_{\geq 0}^{r \times n}$ is the common dictionary matrix. 
An illustration of the decomposition is given in \cref{fig:nmf-illus}. 

\begin{figure}[h!]
\tikzstyle{orig}=[draw,thick,minimum width=2.5cm,minimum height=3cm]
\tikzstyle{repr}=[draw,thick,minimum width=1cm,minimum height=3cm]
\tikzstyle{dict}=[draw,thick,fill=orange!20,minimum width=2.5cm,minimum height=1cm]
\tikzstyle{arrow}=[->,thick]
\centering
\begin{tikzpicture}
\fill[teal!50] (0,0) rectangle (2.5,1.5); 
\fill[teal!20] (0,-1.5) rectangle (2.5,0); 
\draw[orig] (0,-1.5) rectangle (2.5,1.5);
\node at (1.25,0.75) {$\mat X_A$}; 
\node at (1.25,-0.75) {$\mat X_B$};

\fill[magenta!50] (4,0) rectangle (5,1.5); 
\fill[magenta!20] (4,-1.5) rectangle (5,0); 
\draw[repr] (4,-1.5) rectangle (5,1.5);
\node at (4.5,0.75) {$\mat W_A$}; 
\node at (4.5,-0.75) {$\mat W_B$};

\node[dict] (H) at (7,1) {$\mat H$};

\node at (3.25,0.75) {$\approx$};

\end{tikzpicture}
\caption{Illustration of NMF applied to a data matrix $\mat X$ that consists of two submatrices $\mat X_A$ and $\mat X_B$. 
The matrices $\mat W_A$ and $\mat W_B$ are the representation matrices corresponding to $\mat X_A$ and $\mat X_B$, respectively, and $\mat H$ the common dictionary matrix.}
\label{fig:nmf-illus}
\end{figure}

In the standard NMF, defined in \cref{eq:nmf_form}, we have:
\begin{equation*}
\underset{\mat W \in \R_{\geq 0}^{m \times r}, \mat H \in \R_{\geq 0}^{r \times n}}{\argmin}\|\mat X - \mat W \mat H\|^2 = \underset{\substack{\mat W_A \in \R_{\geq 0}^{m_1 \times r}, \mat W_B \in \R_{\geq 0}^{m_2 \times r} \\ \mat H \in \R_{\geq 0}^{r \times n}}}{\argmin} \left( \|\mat X_A - \mat W_A \mat H\|^2 + \|\mat X_B - \mat W_B \mat H\|^2  \right ).
\end{equation*}
Note that both groups are weighted equally and we seek to minimize the sum of the reconstruction error of each group in the joint decomposition.
The problem can be written for $L$ mutually exclusive groups,
\begin{equation}
\label{eq:nmf_split} 
\underset{\mat W \in \R_{\geq 0}^{m \times r}, \mat H \in \R_{\geq 0}^{r \times n}}{\argmin}\|\mat X - \mat W \mat H\|^2 = 
\underset{\substack{\mat W_\ell \in \R_{\geq 0}^{m_\ell \times r}, \forall \ell \in \{1, \cdots, L \} \\ \mat H \in \R_{\geq 0}^{r \times n}}}{\argmin} \quad \sum\limits_{\ell =1}^ L  \|\mat X_\ell - \mat W_\ell \mat H\|^2  .
\end{equation}

This standard objective function is designed to perform well \textit{on average}.
It seeks an overall low reconstruction error which disregards the size and complexity of each data group.
For example, in the case of an imbalanced dataset where $|A| \gg |B|$ an overall low reconstruction error does not guarantee that the reconstruction error restricted to group $B$ is low. 
Additionally, the standard objective function does not pay attention to the complexity of the data groups.

\begin{figure}[h!]
\centering
\includegraphics[height=2.2in]{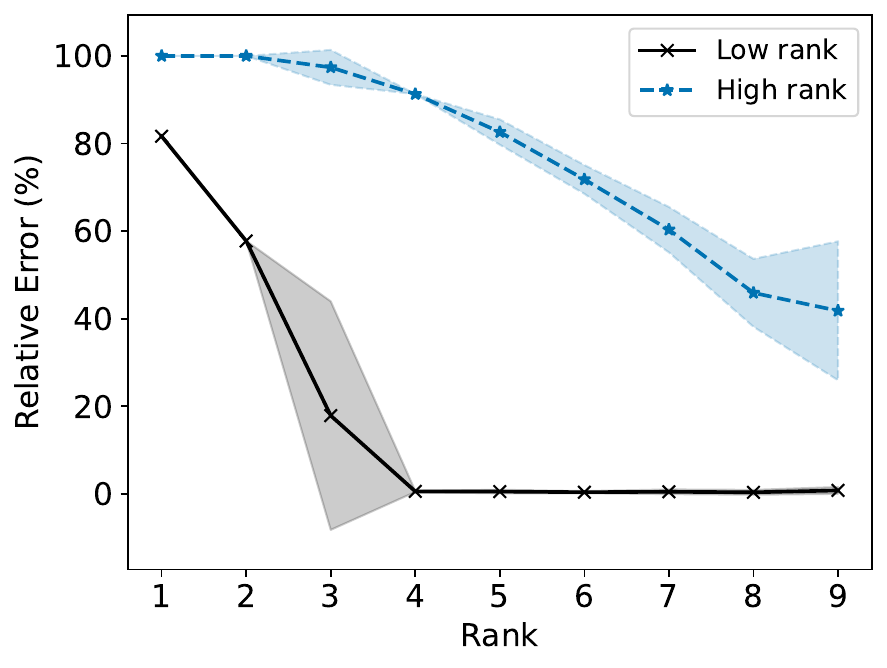}
\includegraphics[height=2.2in]{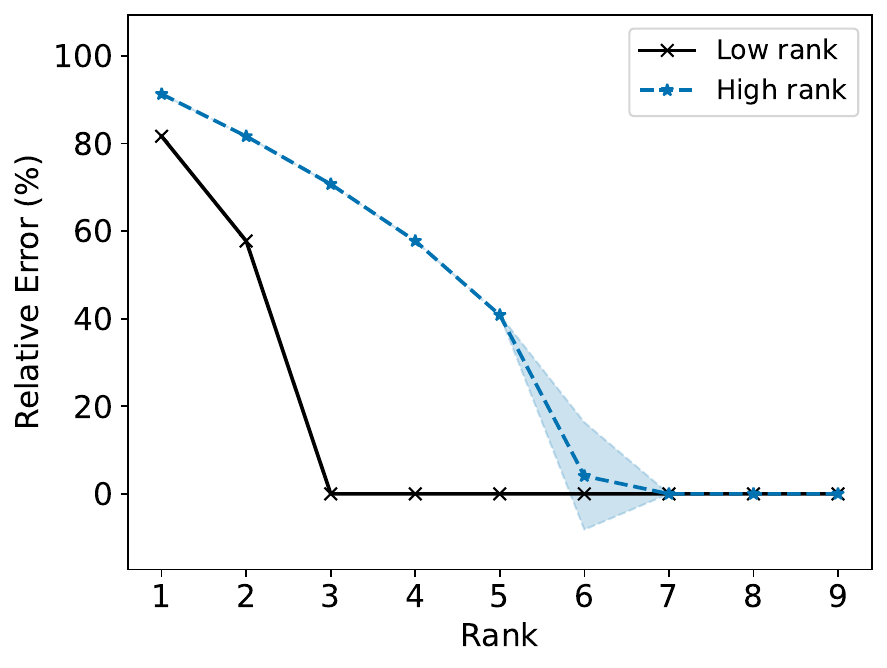}
\caption{
The Relative Error (\%) of each group is reported using
standard NMF applied on the full synthetic data matrix (left) and each group data matrix individually (right).  This synthetic data matrix is composed of two groups in orthogonal subspaces: a high rank ($r=6$) group and a low rank ($r=3$) group.  The mean and standard deviation over 10 trials is reported.
}
\label{fig:snmf_ortho_1_error}
\end{figure}

\begin{figure}[h!]
\centering
\includegraphics[height=2.2in]{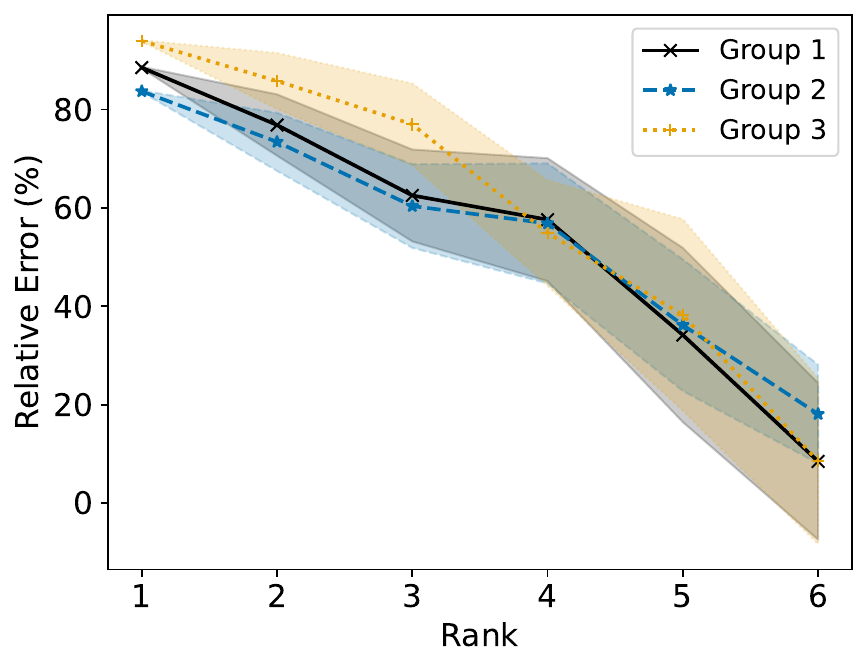}
\includegraphics[height=2.2in]{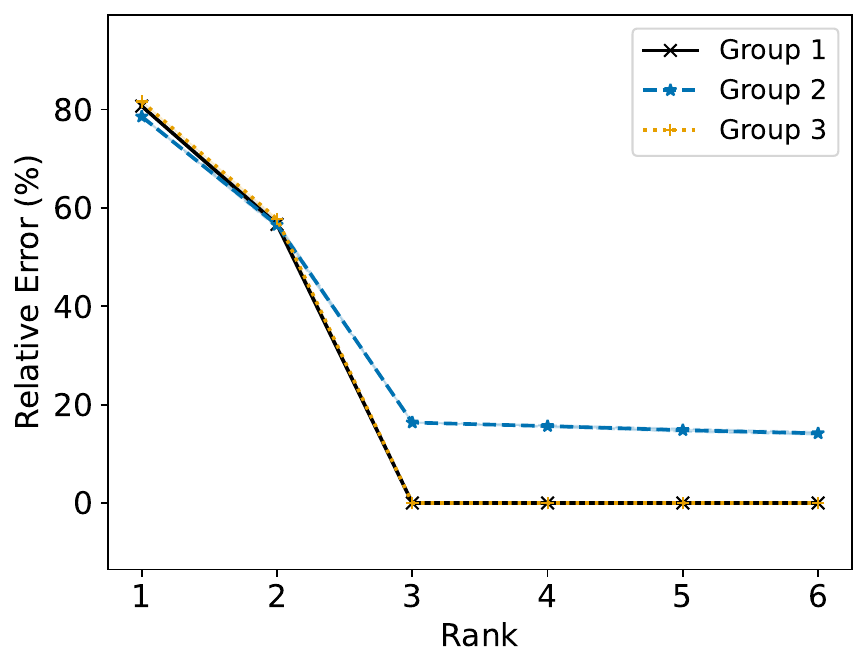}
\caption{
The Relative Error (\%) of each group is reported using
standard NMF applied on the full synthetic data matrix (left) and each group data matrix individually (right).  This synthetic data matrix consists of three groups.  Groups 1 and 3 lie in orthogonal subspaces and are each rank 2.  Group 2 is constructed from the same non-negative basis as group 1, but with random noise added. The mean and standard deviation over 10 trials is reported.}
\label{fig:snmf_ortho_2_error}
\end{figure}

We consider now some illustrative experiments.  \Cref{fig:snmf_ortho_1_error,fig:snmf_ortho_2_error} show two different experiments with synthetic data to highlight two different situations where standard NMF can produce ``unfair" results.  In all of these experiments, there exist multiple groups such that we wish to obtain a factorization that works well for all groups simultaneously.  In \cref{fig:snmf_ortho_1_error}, one of these groups is lower rank than the other.  In this case, we see that the decompositions perform much better on the lower rank group ($r=3$) than the higher rank group ($r=6$).  In particular, the reconstructions for rank $6$ and above have large error for the high rank group, even though this group is rank $6$.  In \cref{fig:snmf_ortho_2_error}, we have three different groups of same size and two of them (groups 1 and 2) lie in approximately the same data subspace.  Here, groups 1 and 2 typically have better reconstruction than group 3 for low rank decompositions. While all groups have the same size and magnitude, sharing a similar basis introduces an imbalance in size (thus magnitude) in the full dataset. This is a also common scenario in real-world settings that is important to consider.  
In both \cref{fig:snmf_ortho_1_error,fig:snmf_ortho_2_error}, we see that good low rank reconstructions of the dataset are possible when standard NMF is applied to each group individually.  The full details of these experiments, including the details for generating the synthetic data, are in \cref{sec:syn_exp}.

Minimizing the maximum \textit{reconstruction error} may seem desirable, but this approach favors optimizing for the group with inherently higher complexity when the group sizes and magnitudes are all the same.  Similarly, this approach favors optimizing for the group with the largest size and magnitude when the rank is the same for all groups.
In the next section, we present a fairness criterion that takes into account the size and complexity of the data groups.

\subsection{\fnmf Objective}
\label{sec:fnmf-objective}

In our consideration of a fairer NMF, we work under a min-max fairness framework with a criterion similar to that of Fair-PCA~\citep{samadi2018price}.
We consider the following definition of \emph{reconstruction loss} of a group.
\begin{definition}[Relative Reconstruction Loss] 
\label{def:rec_loss} 
Suppose we have a group data matrix $\mat X_{\ell} \in \R_{\geq 0}^{m_{\ell} \times n}$, desired rank $r \in \mathbb N$, and $E_\ell$ representing the error obtained by replacing $\mat X_{\ell}$ with a $r$-rank NMF approximation.
For a given $\mat W_\ell \in \mathbb{R}^{m_\ell \times r}$ and $\mat H   \in \R _{\geq 0}^{r \times n}$, we define the relative reconstruction loss of $\mat X_{\ell}$ as,
\[
\frac{\|\mat X_{\ell} - \mat W_{\ell} \mat H \| -E_\ell}{\|\mat X_{\ell}\|}.
\]
\end{definition}

\begin{remark}
\label{rem:rec_loss}
A reasonable choice for $E_\ell$ is to take
\[E_\ell = \mathbb{E}_{\mat X_{\ell}^*}\|\mat X_{\ell} - \mat X_{\ell}^*\|\]
where the expectation is taken over the $\mat X_{\ell}^*$ obtained from a specific randomized implementation of rank-$r$ NMF on $\mat X_\ell$.  
\end{remark}

\begin{remark}
    Note here and throughout that we assume a priori knowledge of the $L$ population subgroups. This allows us to focus on the fairness objective here, and while it is reasonable in many settings, we remark on possible relaxations of this assumption in \cref{sec:discuss}.
\end{remark}

We provide details on the numerical approximation in \cref{sec: fnmf-algorithms}.  We again highlight that finding an optimal non-negative rank-$r$ NMF of a matrix $\mat X_{\ell} \in \R_{\geq 0}^{m_{\ell} \times n}$ is NP-hard~\citep{vavasis2010complexity}. Taking the expectation over a randomized NMF implementation allows $E_\ell$ to compensate both for the underlying dimensionality of the group and for how difficult it is for the NMF algorithm to find a good representation for the group.

Roughly speaking, the reconstruction loss is the difference between how well group $\mat X_\ell$ is reconstructed via a standard NMF model trained on the entire data $\mat X$ and a standard NMF model trained on $\mat X_\ell$ only.
We further note that we normalize by the Frobenius norm of the group matrix to mitigate for the size of the group and the actual varying magnitudes of $\| \mat X _ \ell\|$.

We emphasize that our goal is to learn a common non-negative low-rank NMF model for all groups, rather than separate NMF models for each group.
In \fnmf, we seek to minimize the maximum of the average reconstruction loss across $L$ different groups:

\begin{equation}
\label{eq:fnmf}
\underset{\substack{\mat W \in \R _ {\geq 0}^{m \times r} \mat H   \in \R _ {\geq 0}^{r \times n}}}{\min} \quad  \max \limits _{\ell \in \{1, \cdots, L \}}  \left \{\frac{\|\mat X _\ell - \mat W_ \ell \mat H\| - E_\ell}{\| \mat X   _\ell \|}\right \}
\end{equation}

First, we note that the second term in the objective function $E_\ell$ is a fixed pre-computed constant for a given group $\ell$ as defined in \cref{rem:rec_loss}.
This optimization problem seeks to learn a common NMF model that minimizes the maximum deviation from the group-wise ``optimal'' NMF approximation across all groups.

Second, we note that the loss function in \cref{eq:fnmf} is incomplete.  We wish to select $\mat H$ that minimizes the loss:
\begin{equation*}
\max \limits _{\ell \in \{1, \cdots, L \}}  \left \{\frac{\|\mat X _\ell - \mat W_ \ell \mat H\| - E_\ell}{\| \mat X   _\ell \|}\right \}.
\end{equation*}
However, in general, $\mat W$ is under-specified.
We can freely perturb $\mat W_\ell$ for any group that does not attain the maximum loss and achieves another solution of \cref{eq:fnmf}, provided the loss for group $\ell$ does not grow too large.  This is because the group representation matrices $\mat W_\ell$ all act independently of each other.  To resolve this issue, we choose
\begin{equation*}
    \min\limits_{\mat W _\ell \in \R _ {\geq 0}^{m_\ell \times r}}\|\mat X_\ell - \mat W_\ell \mat H\|
\end{equation*}
for all $\ell$ independently of each other, which is equivalent to choosing
\begin{equation}\label{eq:grp_error}
    \min\limits_{\mat W \in \R _ {\geq 0}^{m \times r}}\|\mat X - \mat W \mat H\|.
\end{equation}
We remark the optimal $\mat W$ for a fixed $\mat H$ may give different losses for different groups even if $\mat H$ is the exact minimum given by \cref{eq:fnmf}.  While this may result in one group having a lower loss than another, this inequality is ``free'' in the sense that it does not come at the expense of another group. 
Therefore, in \fnmf we seek to minimize both \cref{eq:fnmf} and \cref{eq:grp_error}, prioritizing the first over the second.

Referring back to~\cref{sec:related-works-unsupervised}, \citet{buet2022towards} consider fairness criteria that aim to reduce disparity across group-wise average costs through a penalized learning approach.
Their disparity penalty is defined as the difference between each group's average loss and the overall average loss across all groups -- a constant term shared by all groups.
In contrast, we formulate fairness as minimizing the maximum group-wise disparity, where each group's performance is compared against the standard NMF objective evaluated on that group alone.
This distinction leads to different fairness criteria and trade-offs.
Additionally, we solve a constrained optimization problem with hard non-negativity constraints, whereas~\citet{buet2022towards} incorporate non-negativity through penalty terms in an unconstrained formulation.

\section{Algorithms}
\label{sec: fnmf-algorithms}

We present two algorithms for solving the \fnmf problem formulation.

\subsection{Estimating \texorpdfstring{$E_\ell$}{E\_l}}\label{sec:estimating_E-ell}

As discussed in \cref{rem:rec_loss}, for a group $\ell$, a reasonable choice for the estimate of the optimal rank-$r$ error, $E_\ell$, is to take the expectation over the error obtained from a specific randomized implementation of rank-$r$ NMF for the group.  This leads to a natural algorithm for estimating $E_\ell$. It suffices to sample the single group NMF reconstruction $T$ times and take the average.  This is described in \cref{alg:E-ell}.

A consequence of this choice in algorithm is that the relative reconstruction loss for a group may be negative.  Using a specific randomized algorithm for NMF and taking the expectation of the reconstruction error provides an upper bound for the group reconstruction error, not a lower bound.  When a factorization for a group is then obtained using a different algorithm, such as \fnmf, there is no reason in general to expect this error to always be worse than our estimate for $E_\ell$.  While counterintuitive, this is not a problem for the \fnmf algorithm.  By directly choosing $E_\ell$ using reconstructions for just the single group, a negative loss will only occur when the presence of other groups does not degrade the reconstruction for a single group.

\begin{algorithm}
\caption{Estimating $E_\ell$}
\label{alg:E-ell}
\begin{algorithmic}
\INPUT{data matrix $\mat X_\ell \in \R^{m_\ell \times n}_{\geq 0}$ of group $\ell$; desired dimension $r \in \mathbb N$; number of estimates $T \geq 1$}
\State $E \gets 0$
\For{$t \gets 1\text{ to }T$}
\State $ \mat W _\ell,\mat H  _\ell \gets \argmin \limits _{\substack{\mat W \in\R^{m_\ell \times r}_{\geq 0}, \mat H  \in\R^{r\times n}_{\geq 0}}} \|\mat X   _\ell - \mat W \mat H  \|$ \Comment{Approximate using standard NMF}
\State $E \gets E + \|\mat X_{\ell} - \mat W _\ell \mat H  _\ell\|$
\EndFor
\State $E_\ell \gets E / T$
\OUTPUT{rank-$r$ error $E_\ell$}
\end{algorithmic}
\end{algorithm}

\subsection{Alternating Minimization 
(AM) Scheme}
The optimization problem in \cref{eq:fnmf} is non-convex with respect to $\mat H$ and $\mat W _\ell$ for all $\ell$. 
However, it is convex with respect to one of the factor matrices while all others are held fixed. 
Further, the corresponding constraint sets are convex.
This allows us to solve the problem using an AM approach on a multi-convex problem as outlined in \cref{alg:AM-fairnmf}.
The AM scheme solves a convex problem in each minimization step, ensuring that there is a global minimum.
Solving for $\mat H ^{(k)}$ and $\mat W ^{(k)}$ in \cref{alg:AM-fairnmf} is similar to solving a standard NMF problem using the AM approach.

\begin{algorithm}
\caption{\fnmf: Alternating Minimization Scheme}
\label{alg:AM-fairnmf}
\begin{algorithmic}
\INPUT{data matrix $\mat X   = 
\begin{bmatrix} 
\mat X   _1 \in \R^{m_1 \times n}_{\geq 0}\\
\vdots\\
\mat X   _L \in \R^{m_L \times n}_{\geq 0}
\end{bmatrix}$ with $L$ groups; desired dimension $r \in \mathbb N$}
\State Compute $E_\ell$ for each group $\ell$ \Comment{e.g., through \cref{alg:E-ell}}
\State Randomly initialize 
$\mat W ^{(0)}= 
\begin{bmatrix} 
\mat W_1^{(0)} \in \R^{m_1 \times r}_{\geq 0}\\
\vdots\\
\mat W_L^{(0)} \in \R^{m_L \times r}_{\geq 0}
\end{bmatrix}$
\State $k \gets 0$
\While{not converged} 
\State $k \gets k + 1$
\State $ \mat H^{(k)} \gets \argmin \limits _{\mat H  \in\R^{r\times d}_{\geq 0}} \quad \max \limits _{\ell \in \{1,\ldots,L\}}\frac{\|\mat X   _\ell - \mat W _\ell^{(k-1)}\mat H  \| - E_\ell}{\|\mat X   _\ell\|}$
\State $ \mat W ^{(k)}= 
\begin{bmatrix} 
\mat W_1^{(k)}\\
\vdots\\
\mat W_L^{(k)}
\end{bmatrix} \gets \argmin \limits _{\mat W \in\R^{n\times r}_{\geq 0}} \|\mat X    - \mat W \mat H  ^{(k)}\|$
\EndWhile
\OUTPUT{coefficient matrix $\mat W ^{(k)}=
\begin{bmatrix} 
\mat W _1^{(k)}\in\R^{m_1 \times r}_{\geq 0}\\
\vdots\\
\mat W _L^{(k)}\in\R^{m_L \times r}_{\geq 0}
\end{bmatrix}$; dictionary matrix $\mat H  ^{(k)} \in \R^{r \times n}_{\geq 0}$}
\end{algorithmic}
\end{algorithm}

As remarked in \cref{eq:nmf_split}, the update rule of $\mat W ^{(k)}$ in \cref{alg:AM-fairnmf} is equivalent to updating the representation matrix of each group $\ell$ as
\[ \mat W _\ell^{(k)} = \argmin \limits _{\mat W_\ell \in\R^{n\times r}_{\geq 0}} \|\mat X_\ell    - \mat W_\ell \mat H  ^{(k)}\|. \]
Consider the function $f$ defined as
\begin{equation}
\label{eq:fnmf-max}
f(\mat W^{(k)}, \mat H^{(k)}) = \max \limits _{\ell \in \{1,\ldots,L\}}\frac{\|\mat X   _\ell - \mat W _\ell^{(k)}\mat H^{(k)}  \| - E_\ell}{\|\mat X   _\ell\|},
\end{equation}
where $\mat H^{(k)}$ and $\mat W^{(k)}$ defined in \cref{alg:AM-fairnmf}.
Indeed, 
\[ \mat H  ^{(k)} \gets \argmin \limits _{\mat H  \in\R^{r\times d}_{\geq 0}} f(\mat W^{(k)}, \mat H)\] is a convex optimization problem.
Then, we have that the loss function $f$ is non-increasing $f(\mat W^{(k)}, \mat H^{(k)}) \leq f(\mat W^{(k-1)}, \mat H^{(k-1)})$ where equality is achieved at a stationary point.
Thus, by iterating the updates of $\mat H^{(k)}$ and $\mat W^{(k)}$, we obtain a sequence of estimates whose loss values converge. 

The two optimization functions in \cref{alg:AM-fairnmf} both fall under restricted classes of convex programs that admit specialized solvers.  The problem
\[\min \limits _{\mat H  \in\R^{r\times d}_{\geq 0}} \quad \max \limits _{\ell \in \{1,\ldots,L\}}\frac{\|\mat X   _\ell - \mat W _\ell^{(k-1)}\mat H  \| - E_\ell}{\|\mat X   _\ell\|}\]
is equivalent to
\begin{align*}\min \limits _{\substack{\mat H \in\R^{r\times d}_{\geq 0} \\t \in \R}}\quad & t \\
\text{subj. to}\quad & \|\mat X   _\ell - \mat W _\ell^{(k-1)}\mat H  \| \leq t\|\mat X_\ell\| + E_\ell \quad  \quad \forall \ell  \in \{1,\ldots,L\},
\end{align*}
which is a second-order cone program (SOCP).  The minimization problem for $W$ is equivalent to
\[\min \limits _{\mat W \in\R^{n\times r}_{\geq 0}} \|\mat X    - \mat W \mat H  ^{(k)}\|^2,\]
which is a non-negative least squares (NNLS) problem, a specific type of quadratic program (QP).

\subsection{Multiplicative Updates (MU) Scheme}
\label{sec:mu-scheme}

\begin{algorithm}
\caption{\fnmf: Multiplicative Updates Scheme}
\label{alg:MU-fairnmf}
\begin{algorithmic}
\INPUT{data matrix $\mat X   = 
\begin{bmatrix} 
\mat X   _1 \in \R^{m_1 \times n}_{\geq 0}\\
\vdots\\
\mat X   _L \in \R^{m_L \times n}_{\geq 0}
\end{bmatrix}$ with $L$ groups; desired dimension $r \in \mathbb N$}
\State Compute $E_\ell$ for each group $\ell$ \Comment{e.g., through \cref{alg:E-ell}}
\State Randomly initialize 
$\mat H  ^{(0)} \in \R^{r\times n}_{\geq 0}$,
$\mat W ^{(0)}= 
\begin{bmatrix} 
\mat W_1^{(0)} \in \R^{m_1 \times r}_{\geq 0}\\
\vdots\\
\mat W_L^{(0)} \in \R^{m_L \times r}_{\geq 0}
\end{bmatrix}$
\State Initialize $\bm  c^{(0)} = \bm  0 \in \R^L$
\State $k \gets 0$
\While{not converged}
\State $k \gets k + 1$
\State $\displaystyle\ell_* \gets \argmax \limits _{\ell \in \{1,\ldots,L\}}\frac{\|\mat X   _\ell - \mat W _\ell^{(k-1)}\mat H^{(k-1)}  \| - E_\ell}{\|\mat X   _\ell\|}$
\State $\bm  {c}^{(k)} \gets \bm  {c}^{(k-1)} + \bm  {e}_{\ell_*}$\Comment{$\bm  {e}_{\ell_*}$ is $\ell_*$-th standard basis vector}\vspace{3pt}
\State $\displaystyle\mat{\tilde{X}} \gets \begin{bmatrix} 
\mat c^{(k)}_1 \mat X_1 / \|\mat X_1\| \in \R^{m_1 \times n}_{\geq 0}\\
\vdots\\
\mat c^{(k)}_L \mat X_L / \|\mat X_L\|\in \R^{m_L \times n}_{\geq 0}
\end{bmatrix}, \mat{\tilde{W}} \gets \begin{bmatrix} 
\mat c^{(k)}_1 \mat W_1^{(k-1)}/ \|\mat X_1\| \in \R^{m_1 \times r}_{\geq 0}\\
\vdots\\
\mat c^{(k)}_L \mat W_L^{(k-1)}/ \|\mat X_L\| \in \R^{m_L \times r}_{\geq 0}
\end{bmatrix}$\vspace{3pt}
\State $\displaystyle \mat H^{(k)} \gets \mat H^{(k-1)} \odot \frac{\mat {\tilde{W}}^\top \mat{\tilde{X}}}{\mat {\tilde{W}}^\top \mat {\tilde{W}} \mat H^{(k-1)}} $\vspace{3pt}
\State $\displaystyle\mat W ^{(k)}= 
\begin{bmatrix} 
\mat W _1^{(k)}\\
\vdots\\
\mat W_L^{(k)}
\end{bmatrix} \gets \mat{W}^{(k-1)} \odot \frac{\mat{X} \mat{H}^{(k)^\top}}{\mat{W}^{(k-1)}  \mat H^{(k)} \mat H^{(k)^\top}}$
\EndWhile
\OUTPUT{coefficient matrix $\mat W ^{(k)}=
\begin{bmatrix} 
\mat W _1^{(k)}\in\R^{m_1 \times r}_{\geq 0}\\
\vdots\\
\mat W _L^{(k)}\in\R^{m_L \times r}_{\geq 0}
\end{bmatrix}$; dictionary matrix $\mat H  ^{(k)} \in \R^{r \times n}_{\geq 0}$} 
\end{algorithmic}
\end{algorithm}

\begin{figure}[h!]
\centering
\includegraphics[height=2.2in]{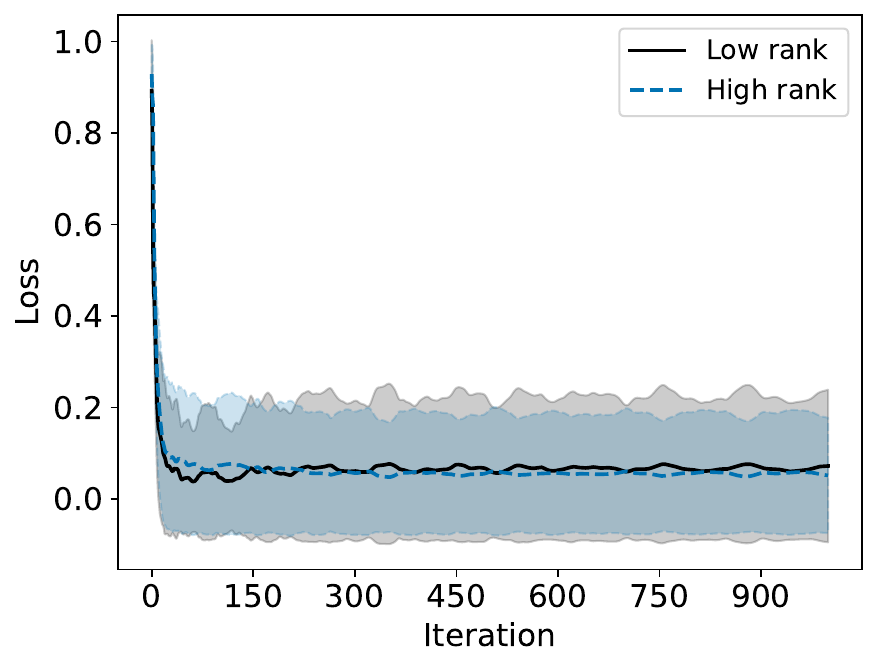}
\includegraphics[height=2.2in]{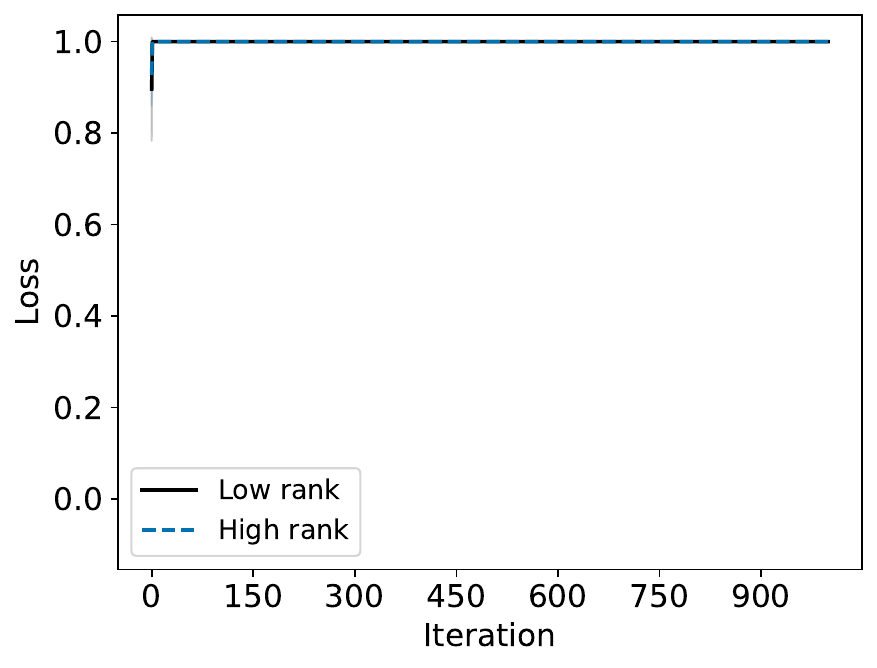}
\caption{
Convergence of the multiplicative update rule for \fnmf on a synthetic data matrix with two groups.   The multiplicative update rule is compared when using the decreasing step size for $\bm{c}$ (left) and when exactly optimizing $\bm{c}$ (right).  The relative reconstruction loss of each group is reported per iteration with the mean and standard deviation taken over $100$ trials.
}
\label{fig:ortho_1_convergence}
\end{figure}

\begin{figure}[h!]
\centering
\includegraphics[height=2.2in]{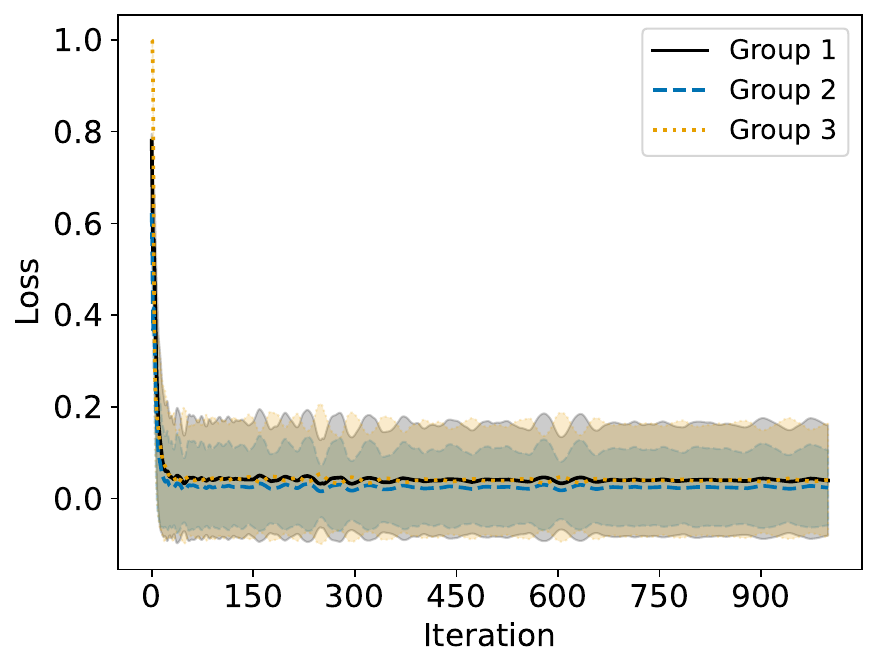}
\includegraphics[height=2.2in]{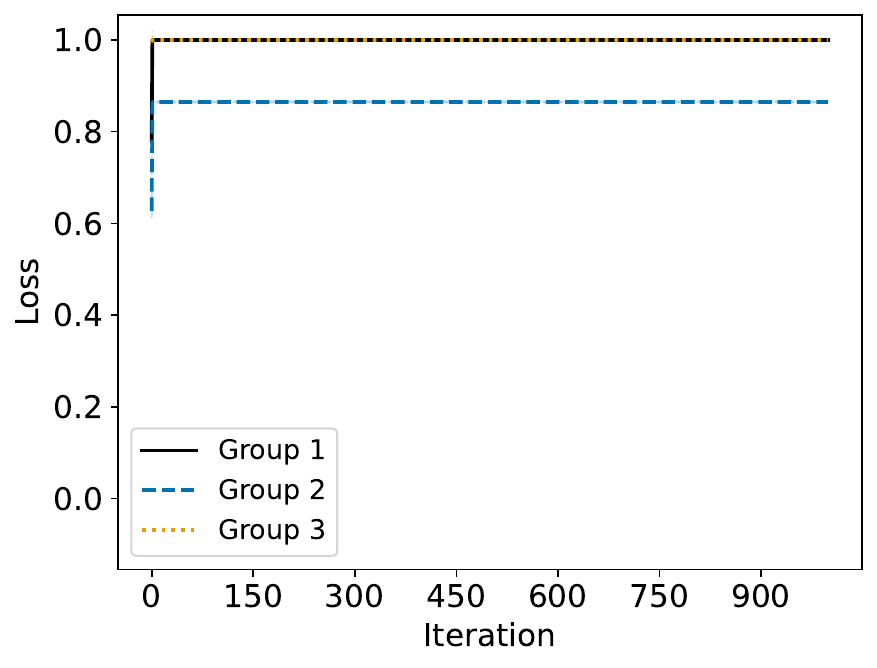}
\caption{
Convergence of the multiplicative update rule for \fnmf on a synthetic data matrix with three groups.   The multiplicative update rule is compared when using the decreasing step size for $\bm{c}$ (left) and when exactly optimizing $\bm{c}$ (right).  The relative reconstruction loss of each group is reported per iteration with the mean and standard deviation taken over $100$ trials.
}
\label{fig:ortho_2_convergence}
\end{figure}

In addition to the AM scheme, we also adapt the multiplicative updates (MU) scheme for the \fnmf problem formulation.  Consider an equivalent form to the loss function in \cref{eq:fnmf-max}:
\[f(\mat W^{(k)}, \mat H^{(k)}) = \max \limits _{\substack{\bm {c} \in \R^L \\\|\bm {c}\|_1 = 1}} g(\mat W^{(k)}, \mat H^{(k)}, \bm {c}) := \sum_{\ell=1}^L \bm {c}_\ell\frac{\|\mat X   _\ell - \mat W _\ell^{(k)}\mat H^{(k)}  \| - E_\ell}{\|\mat X   _\ell\|}.\]
Let $\bm {c}^{(k)}$ be our estimate of the maximizer $\bm  c$ and consider an alternating approach by maximizing the function $g(\mat {W}, \mat H,\bm {c})$ in $\bm {c}$ and minimizing in $\mat{W}$ and $\mat{H}$.  The maximizer $\bm {c}^{(k)}$ at the $k$-th iteration is simply $\bm {e}_{\ell_*}$ (the $\ell_*$-th standard basis vector) where,
\[\ell_* = \argmax\limits_{\ell\in\{1,\ldots,L\}}\frac{\|\mat X_\ell - \mat W _\ell^{(k-1)}\mat H^{(k-1)}  \| - E_\ell}{\|\mat X   _\ell\|}.\]
Start with $\bm {c}^{(0)} = \bm {0}$.  By setting $\bm {c}^{(k)} = \frac{k-1}{k}\bm {c}^{(k-1)} + \frac{1}{k}\bm {e}_{\ell_*}$ we update with a decreasing step size while ensuring $\|\bm {c}^{(k)}\|_1 = 1$.  This is desirable as just setting $\bm {c}^{(k)} = \bm {e}_{\ell_*}$ can result in too much oscillation in the largest loss group and therefore poor convergence.  This is demonstrated in \cref{fig:ortho_1_convergence,fig:ortho_2_convergence}, using the synthetic data discussed in detail in \cref{sec:syn_exp}.  Exactly optimizing $\bm{c}$ results in an algorithm which does not converge to a low loss solution.

For fixed $\bm {c}^{(k)}$ and $\mat{W}^{(k-1)}$, we minimize $g$ in $\mat{H}$ by selecting
\begin{equation}\label{eq:mu_min_H_real}\argmin\limits_{\mat{H}\in\R^{r\times d}_{\geq 0}} \sum_{\ell=1}^L \bm {c}_\ell^{(k)}\frac{\|\mat X_\ell - \mat W_\ell^{(k-1)} \mat H\|}{\|\mat X_\ell\|}.
\end{equation}
A related but easier problem to solve is
\begin{equation}\label{eq:mu_min_H_alt}\argmin\limits_{\mat{H}\in\R^{r\times d}_{\geq 0}} \sum_{\ell=1}^L \left(\bm {c}_\ell^{(k)}\right)^2\frac{\|\mat X_\ell - \mat W_\ell^{(k-1)} \mat H\|^2}{\|\mat X_\ell\|^2} = \argmin\limits_{\mat{H}\in\R^{r\times d}_{\geq 0}}\|\mat{\tilde{X}} - \mat{\tilde{W}} \mat H\|^2
\end{equation}
for the following two block matrices:
\[ \mat{\tilde{X}} = \begin{bmatrix} 
\mat c^{(k)}_1 \mat X_1 / \|\mat X_1\| \in \R^{m_1 \times n}_{\geq 0}\\
\vdots\\
\mat c^{(k)}_L \mat X_L / \|\mat X_L\|\in \R^{m_L \times n}_{\geq 0}
\end{bmatrix}, \mat{\tilde{W}} = \begin{bmatrix} 
\mat c^{(k)}_1 \mat W_1^{(k-1)}/ \|\mat X_1\| \in \R^{m_1 \times r}_{\geq 0}\\
\vdots\\
\mat c^{(k)}_L \mat W_L^{(k-1)}/ \|\mat X_L\| \in \R^{m_L \times r}_{\geq 0}
\end{bmatrix}.\]
The solution of \cref{eq:mu_min_H_alt} attains a value of $g(\mat W^{(k)},\mat H, \mat c^{(k)})$ that is at most $\sqrt{L}$ that of \cref{eq:mu_min_H_real} that can be considered as a good choice of minimizer.  
Thus, we can select $\mat{H}^{(k)}$ according to the standard multiplicative update for $\mat{\tilde{X}}$ with $\mat{\tilde{W}}$.  Once we've obtained $\mat H^{(k)}$, we can use the multiplicative update for the single group NMF problem $\mat X_\ell \approx \mat W_\ell \mat H$ to obtain $\mat W_\ell^{(k)}$ for each group $\ell$.  This procedure is described in \cref{alg:MU-fairnmf}.

We make two concluding remarks about this algorithm:
\begin{enumerate}
    \item Enforcing $\|\bm {c}^{(k)}\|_1 = 1$ is actually unnecessary, as the multiplicative update is invariant under scalar multiplication for $\mat{\tilde{X}}$ and $\mat{\tilde{W}}$.  Therefore it suffices to just set $\bm {c}^{(k)} = \bm {c}^{(k-1)} + \bm {e}_{\ell_*}$.
    \item While a learning rate is empirically important for convergence, the exact rate chosen was selected for the simplicity of its implementation and to avoid introducing a hyperparameter which needs to be tuned.  It is likely that other update rules would yield similarly good performance.
\end{enumerate}

\section{Numerical Experiments}
\label{sec:num-exp}
In this section, we evaluate the performance of the AM scheme (\cref{alg:AM-fairnmf}) and the MU scheme (\cref{alg:MU-fairnmf}) for \fnmf.
We compare them with standard NMF on both synthetic benchmark datasets and real-world datasets, including survey and text data, which are common use cases for NMF.
Our code is publicly available\footnote{\url{https://github.com/ErinGeorge/Fairer-NMF}}.

\subsection{Reporting Metrics and Implementations}
\label{sec:implementation}
For all the experiments, we compute the relative reconstruction error given in \cref{def:rec_error} and the relative reconstruction loss given in \cref{def:rec_loss}.
In the figures, “Relative Error (\%)” is the relative reconstruction error scaled by $100$.
We report the mean and standard deviation (given by the shaded region around the mean) over $10$ trials.
In each trial, we re-initialize the algorithm with a random initialization and estimate $E_\ell$ (defined in \cref{rem:rec_loss}) using \cref{alg:E-ell} with $5$ runs ($T=5$).
We use the acronym R-Error to denote the average relative reconstruction error, and R-Loss for the average relative reconstruction loss.

For the AM scheme implementation, we use the open-source package CVXPY~\citep{cvxpy}, which supports a number of different specialized convex program solvers.  
The specific solvers we use are ECOS~\citep{bib:Domahidi2013ecos} and SCS~\citep{ocpb:16} for the SOCP to find $\mat H$ and OSQP~\citep{osqp} for the QP to find $\mat W$.  For the SOCP, we default to ECOS and switch to SCS only if ECOS fails. 
Failures of ECOS do occur, but only for large problem sizes.

For both AM and MU schemes, we iterate until the change in each group's reconstruction error in a single iteration is no more than $10^{-4}$ times the current reconstruction error.  That is when the following condition is met for all groups $\ell$:
\begin{equation}\label{eq:convergence_condition}
  \frac{\left|\|\mat X_\ell - \mat W^{(k)}_\ell \mat H^{(k)}\|-\|\mat X_\ell - \mat W^{(k-1)}_\ell \mat H^{(k-1)}\| \right|}{\|\mat X_\ell - \mat W^{(k)}_\ell \mat H^{(k)}\|} < 10^{-4}.
\end{equation}
This stopping criterion works when no group is able to be reconstructed exactly.  In the synthetic datasets we describe later in this section, this is not true.  For these datasets, we stop if for every group $\ell$ either \cref{eq:convergence_condition} holds or the R-Error is less than $0.1\%$.  The latter only happens for high ranks in our synthetic datasets, when the data can be perfectly reconstructed with a factorization at a given rank.

For all datasets, before running standard NMF or \fnmf, we normalize the features to have unit $\ell_2$-norm.

\subsection{Synthetic Datasets}\label{sec:syn_exp}
For the synthetic datasets, we generate group $\ell$ data matrix $\mat X_\ell \in \R_{\geq 0}^{500 \times 12}$ as $\mat X_\ell = \mat W_\ell \mat H_\ell $ where $\mat W_\ell \in \R_{\geq 0}^{500 \times r_\ell}$ and $\mat H_\ell \in \R_{\geq 0}^{r_\ell \times 12}$.  
We sample the rows of $\mat W_\ell$ independently and uniformly from the set $\{\bm {e}_1^\top,\ldots,\bm {e}_{r_\ell}^\top\}$ (here each standard basis vector $\bm {e}_i \in \R^{r_\ell}$).  That is, we select the rows of $\mat X_\ell$ to be independently and uniformly randomly chosen rows of $\mat H_\ell$.

\subsubsection{Synethetic dataset 1: Different group ranks}
For the first synthetic dataset, we have two groups and take $r_\ell$ and $\mat H_\ell$ to differ for the two groups:
\begin{itemize}
    \item Group 1 (High rank) : $r_1=6$, $\mat H_1 = \begin{pmatrix}\bm {e}_1 & \bm {e}_2 & \bm {e}_3 & \bm {e}_4 & \bm {e}_5 & \bm {e}_6\end{pmatrix}^\top$. 
    \item Group 2 (Low rank): $r_2=3$, $\mat H_2 = \begin{pmatrix}\bm {e}_7+\bm {e}_{10} & \bm {e}_8 + \bm {e}_{11} & \bm {e}_{9} + \bm {e}_{12} \end{pmatrix}^\top$.
\end{itemize}
Here each standard basis vector $\bm {e}_i \in \R^{12}$.
This construction results in the two groups being orthogonal to each other.

\begin{figure}[h!]
\centering
\includegraphics[height=2.2in]{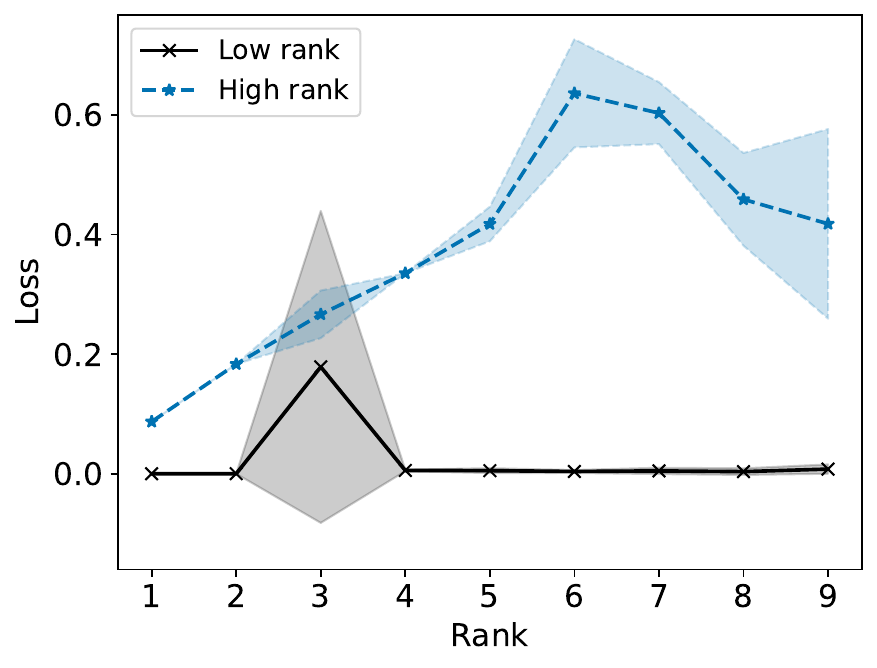}
\caption{Standard NMF applied on the entire first synthetic data matrix. The reconstruction loss of each group is reported for ranks 1 to 9.}
\label{fig:snmf_ortho_1_loss}
\end{figure}

\begin{figure}[h!]
\centering
\includegraphics[height=2.2in]{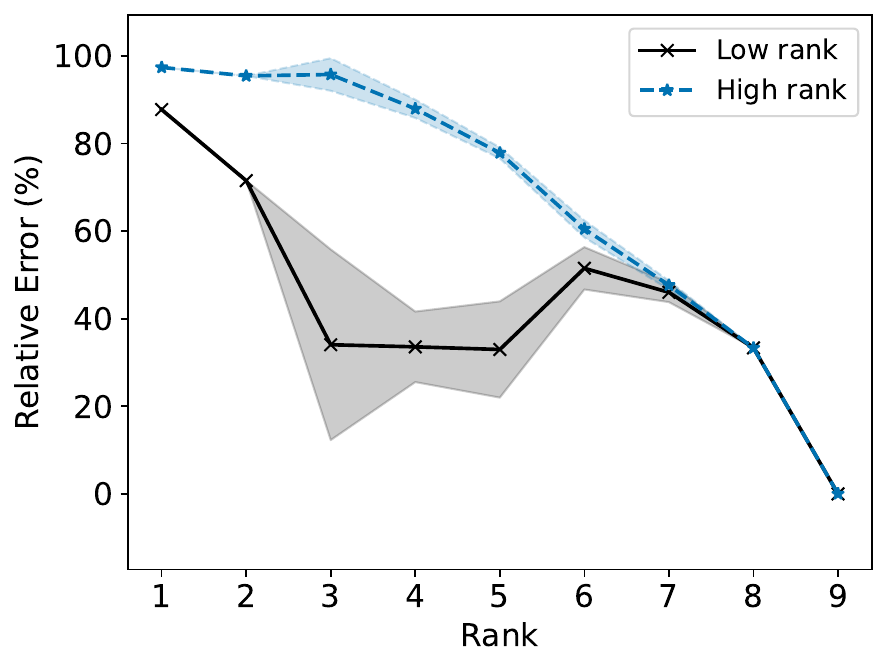} \includegraphics[height=2.2in]{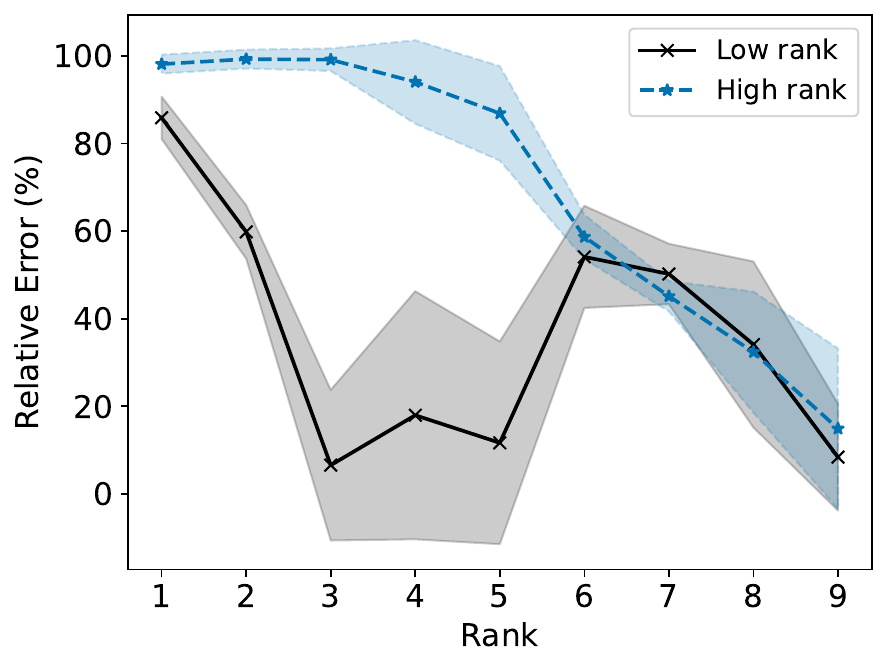}
\caption{
\fnmf applied on the first synthetic data matrix.
The Relative Error (\%) of each group is reported for ranks 1 to 9.
Left: \fnmf with the alternating minimization scheme.
Right: \fnmf with the multiplicative updates scheme.
}
\label{fig:fnmf_ortho_1_error}
\end{figure}

\begin{figure}[h!]
\centering
\includegraphics[height=2.2in]{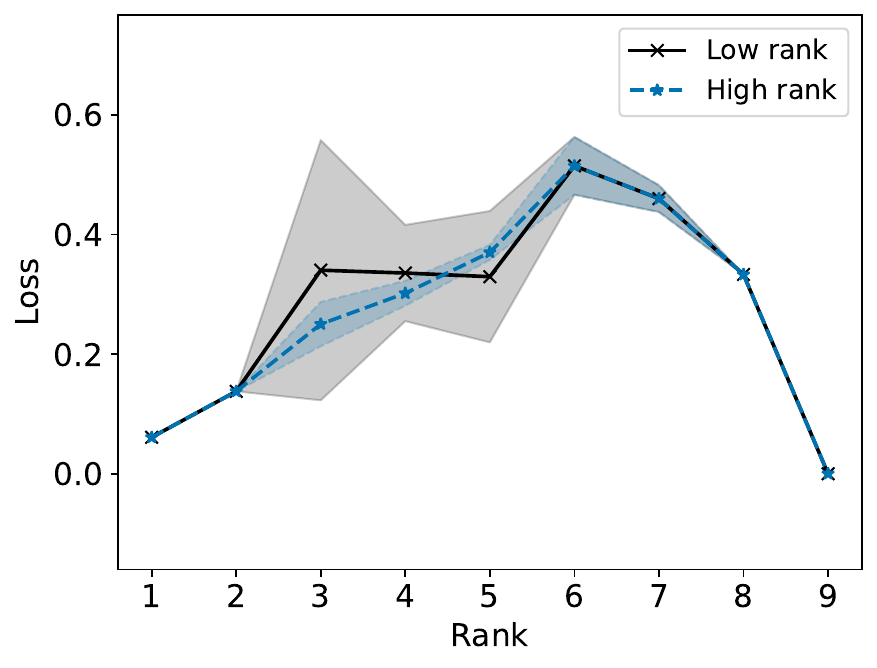} \includegraphics[height=2.2in]{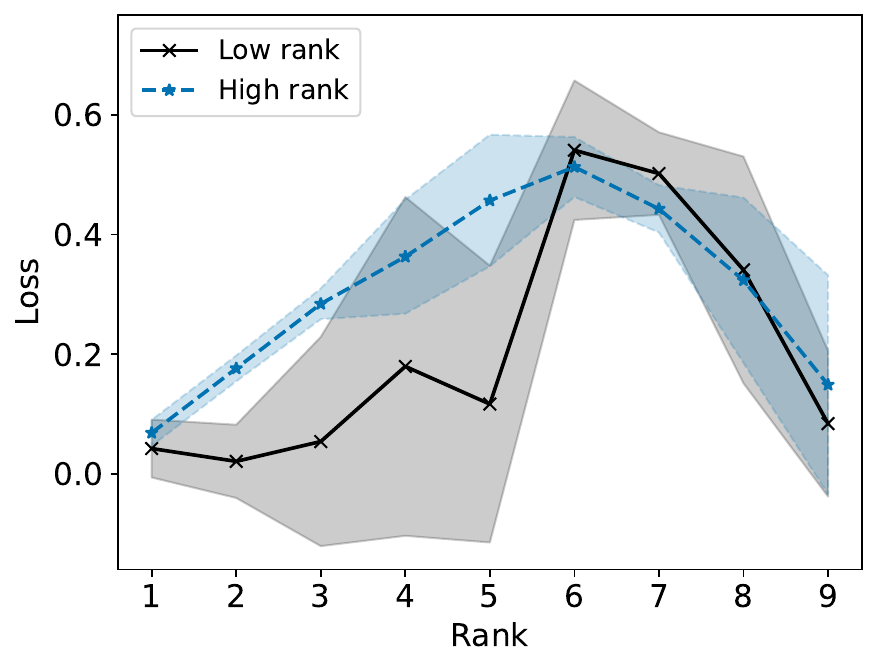}
\caption{
\fnmf applied on the first synthetic data matrix.
The reconstruction loss of each group is reported for ranks 1 to 9.
Left: \fnmf with the alternating minimization scheme.
Right: \fnmf with the multiplicative updates scheme.
}
\label{fig:fnmf_ortho_1_loss}
\end{figure}

As discussed in \cref{sec: nmf-objective}, \cref{fig:snmf_ortho_1_error} shows that standard NMF exhibits a discrepancy in the R-Error among groups with the same size but that differ in complexity.
In \cref{fig:snmf_ortho_1_loss}, for all ranks, we observe a much higher loss for the high rank group ($r=6$) compared to the low rank group $(r=3)$.  This is of course not surprising, given that NMF minimizes total error, which is more efficiently done by minimizing the error of the low rank group.  The loss can be interpreted as the difference between the reconstruction error the group would incur by being part of the population and the error the group would have incurred if the model was run on that group alone.  
Thus, groups with the higher loss are ``sacrificing'' more by being part of the population. 

In \cref{fig:fnmf_ortho_1_error} for \fnmf (MU and AM), we see how the high-rank group ($r=6$) still incur larger R-Error than the low-rank group ($r=3$) for ranks 1 through 5.  Starting from rank 6, however, the \fnmf reconstructions have similar R-Error for both groups. This example, designed to be extreme, also highlights that the fairer formulation has the ability to \textit{increase} error for some individuals or groups.  In fact, this is not surprising, since especially when groups are quite different, the factorization needs more rank to explain all groups, and when that rank is fixed, some groups will have to experience increased error for others to experience decreased error.  The hope of course, is that the former is small for each individual while the latter may improve things significantly for others. If the groups are incredibly different with little in common, one may of course opt to simply treat them as entirely separate populations.  All this being said, as mentioned in the introduction and discussed more in Section \ref{sec:discuss}, what is fair is highly application dependent and algorithm selection should always be done with care.
We also observe in \cref{fig:fnmf_ortho_1_loss} that all groups have a comparable R-Loss, although the MU scheme is less effective at finding a minimum that equalizes the loss between the two groups.

\subsubsection{Synthetic dataset 2: Overlapping subspace structure}
For the second synthetic dataset, we take $r_\ell=3$ for all groups.  We take $\mat H_\ell$ to differ for the three groups:
\begin{itemize}
    \item Groups 1 and 2: $\mat H_1 = \mat H_2  = \begin{pmatrix}\bm {e}_1+\bm {e}_{4} & \bm {e}_2 + \bm {e}_{5} & \bm {e}_{3} + \bm {e}_{6} \end{pmatrix}^\top$.
    \item Group 3: $\mat H_3 = \begin{pmatrix}\bm {e}_7+\bm {e}_{10} & \bm {e}_8 + \bm {e}_{11} & \bm {e}_{9} + \bm {e}_{12} \end{pmatrix}^\top$.
\end{itemize}
For group 2 in this dataset, we then perturb $\mat X_2$ by adding an error term to each coordinate as an independently sampled Gaussian random variable with mean $0$ and variance $1/100$.  We then truncate any negative component of the resulting matrix to $0$.  As a result, group 3 is orthogonal to groups 1 and 2, while groups 1 and 2 have a similar low rank structure.

\begin{figure}[h!]
\centering
\includegraphics[height=2.2in]{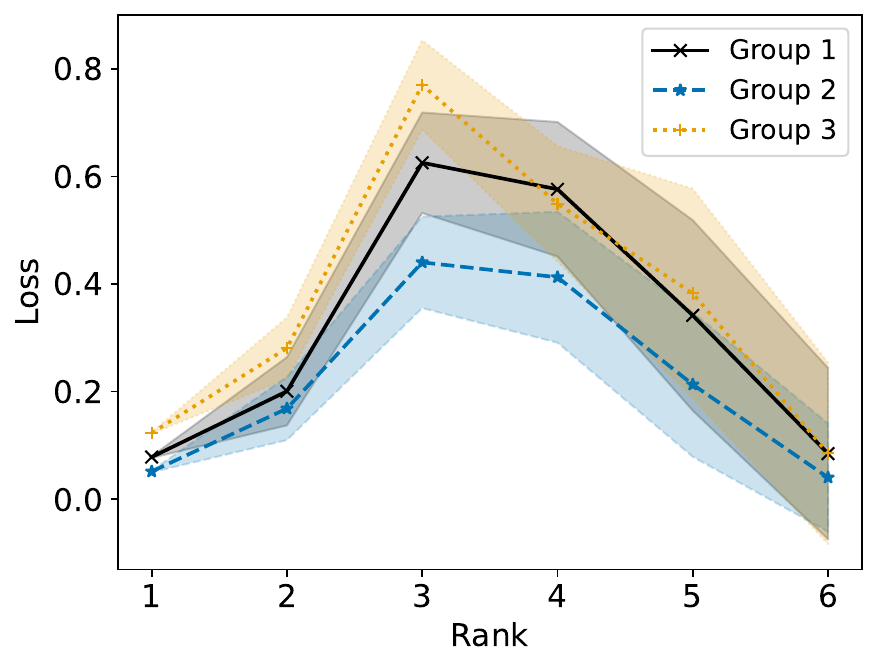}
\caption{Standard NMF applied on the entire second synthetic data matrix. The reconstruction loss of each group is reported for ranks 1 to 6.}
\label{fig:snmf_ortho_2_loss}
\end{figure}

\begin{figure}[h!]
\centering
\includegraphics[height=2.2in]{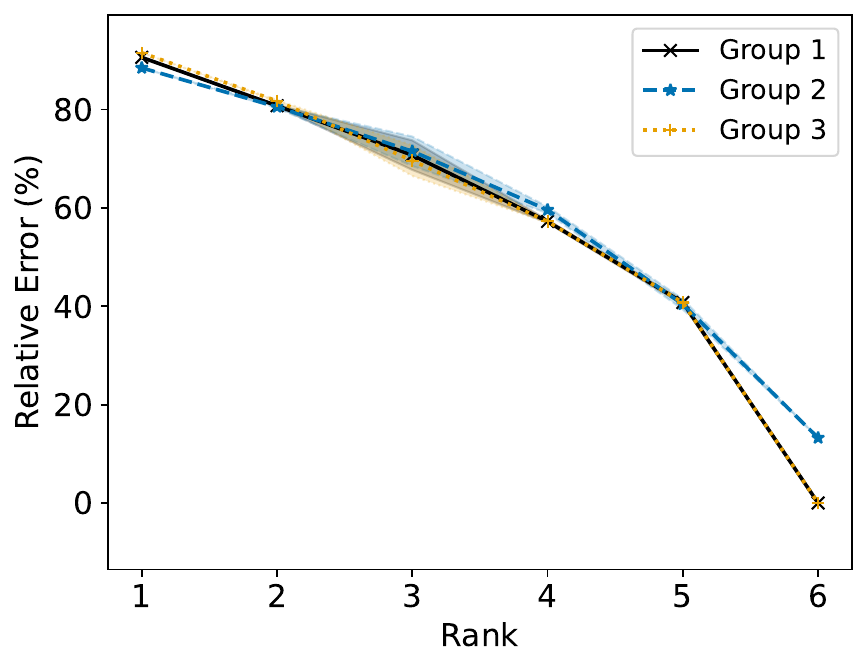} \includegraphics[height=2.2in]{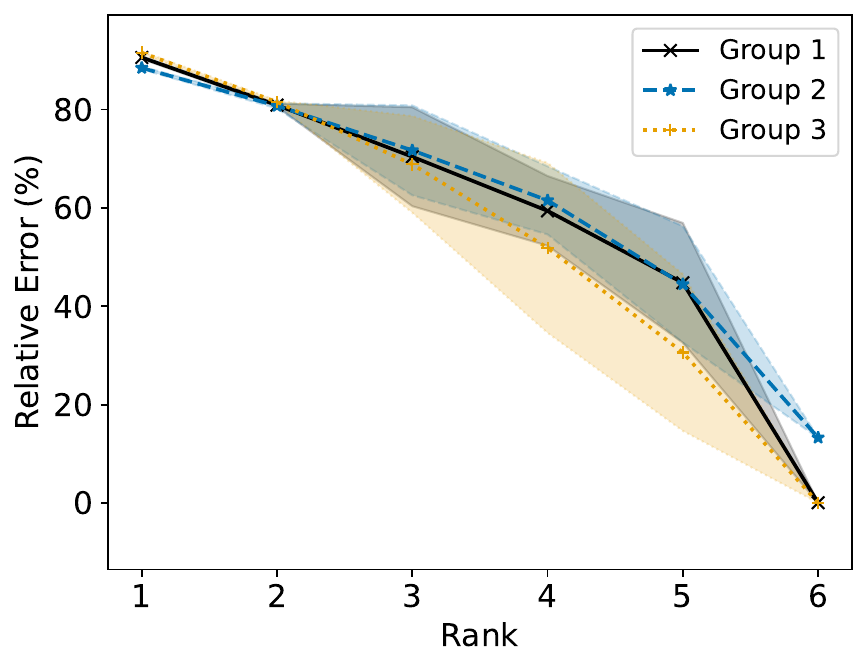}
\caption{
\fnmf applied on the second synthetic data matrix.
The Relative Error (\%) of each group is reported for ranks 1 to 6.
Left: \fnmf with the alternating minimization scheme.
Right: \fnmf with the multiplicative updates scheme.
}
\label{fig:fnmf_ortho_2_error}
\end{figure}

\begin{figure}[h!]
\centering
\includegraphics[height=2.2in]{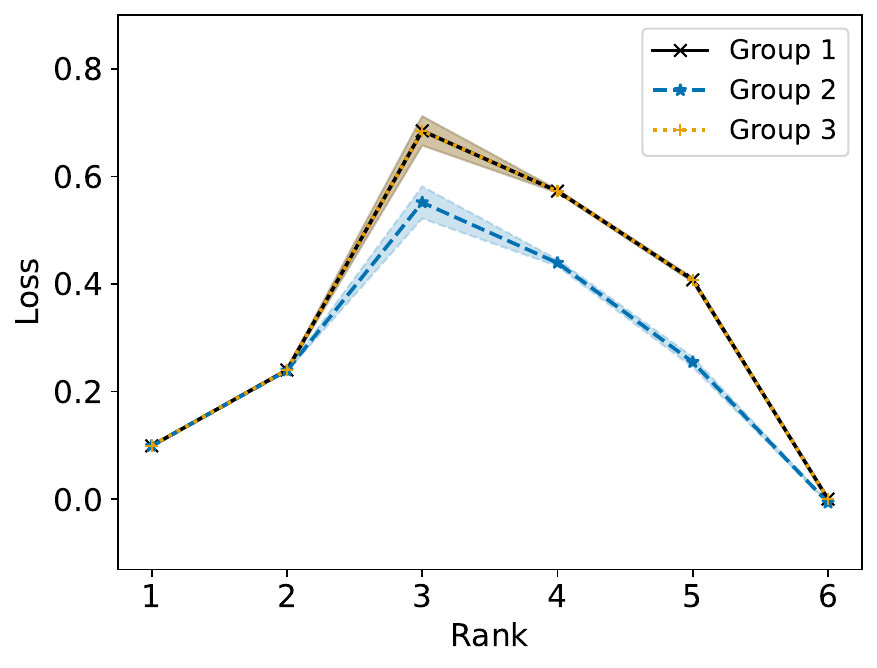} \includegraphics[height=2.2in]{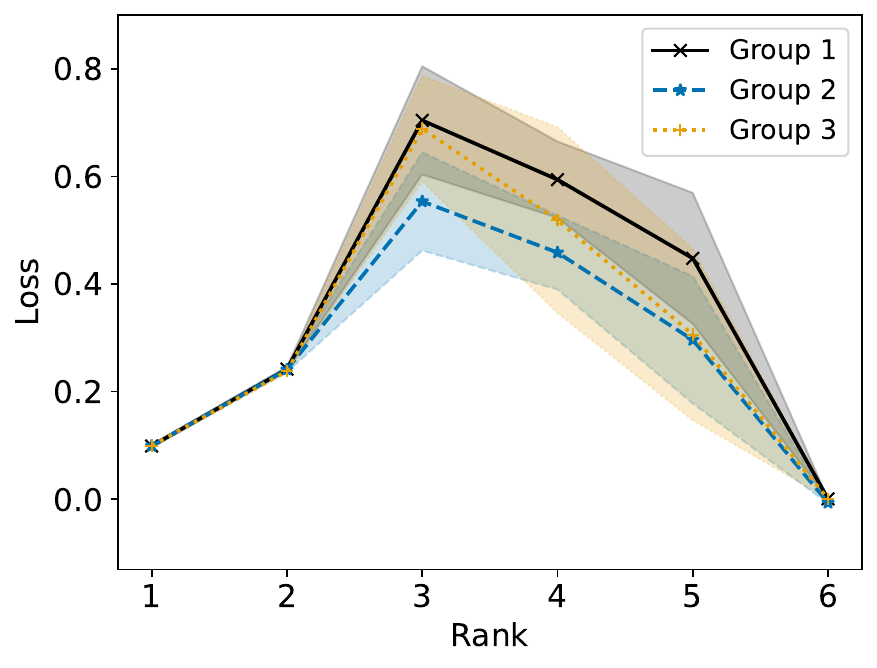}
\caption{
\fnmf applied on the second synthetic data matrix.
The reconstruction loss of each group is reported for ranks 1 to 6.
Left: \fnmf with the alternating minimization scheme.
Right: \fnmf with the multiplicative updates scheme.
}
\label{fig:fnmf_ortho_2_loss}
\end{figure}

In \cref{sec: nmf-objective}, we saw from \cref{fig:snmf_ortho_2_error} that standard NMF with rank at most 3 on the full data matrix results in group 3 having higher R-Error than the other two groups in the second synthetic dataset.  In \cref{fig:snmf_ortho_2_loss}, we see that this is reflected in the loss function, where group 3 has a higher loss for these ranks.  We show the results of \fnmf on this dataset in \cref{fig:fnmf_ortho_2_error,fig:fnmf_ortho_2_loss}.  We can see that \fnmf is able to find a solution where the R-Error for all groups is nearly equal for all ranks in the dataset and where the R-Loss for groups 1 and 2 are equal.  However, the MU scheme again exhibits much higher variance than the AM scheme for this problem.  Furthermore, for the higher rank reconstructions the MU scheme is not able to consistently achieve parity of the R-Error and R-Loss between groups 1 and 3.

\subsection{Heart Disease Dataset}
The heart disease dataset~\citep{heart_disease_45} is a dataset designed for medical research to predict whether a patient has heart disease or not based on various medical attributes.
The dataset is commonly used in machine learning research to evaluate a model's performance in classifying the presence and absence of the disease.
The most complete and commonly used subset of the dataset in machine learning research is the Cleveland database.
The database consists of 303 samples and 13 attributes that are clinical parameters obtained from the patients such as sex, age, resting blood pressure, and serum cholesterol.

\begin{figure}[h!]
\centering
\includegraphics[height=2.2in]{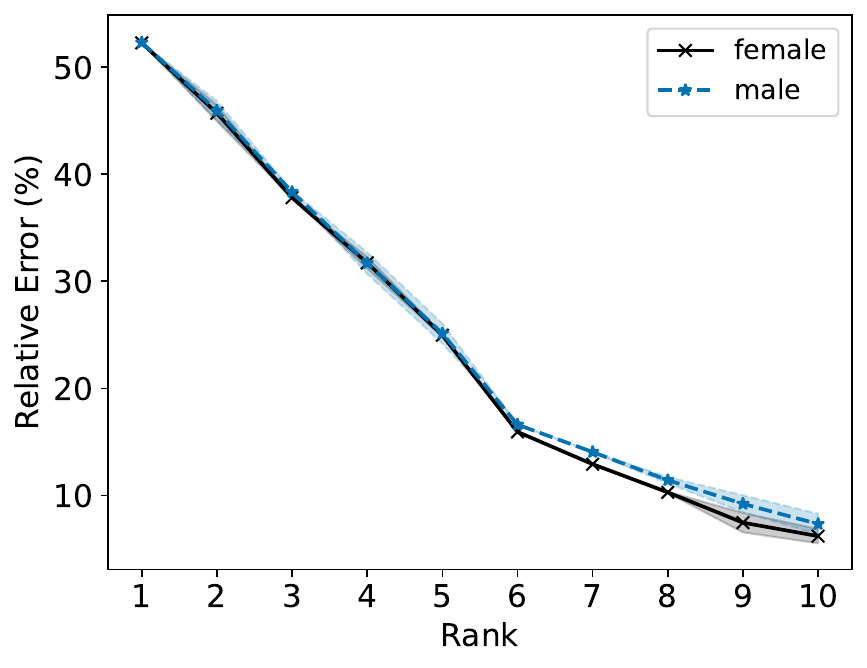} \includegraphics[height=2.2in]{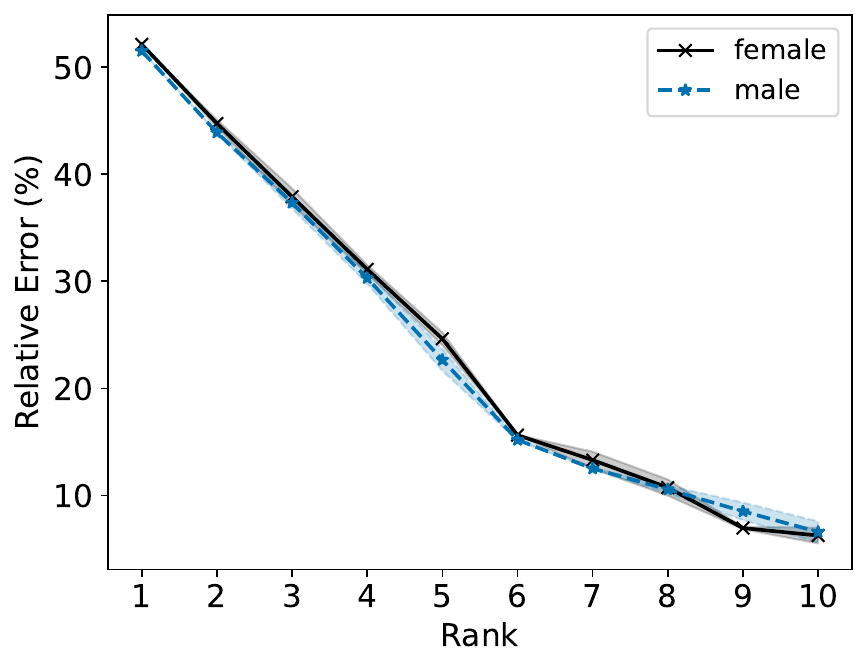}
\caption{
The Relative Error (\%) of each group in the Heart Disease dataset is reported for ranks 1 to 10.
Left: standard NMF applied on the entire Heart Disease data matrix. 
Right: standard NMF applied on the Heart Disease data matrix of each group individually.}

\label{fig:snmf_hd_error}
\end{figure}

\begin{figure}[h!]
\centering
\includegraphics[height=2.2in]{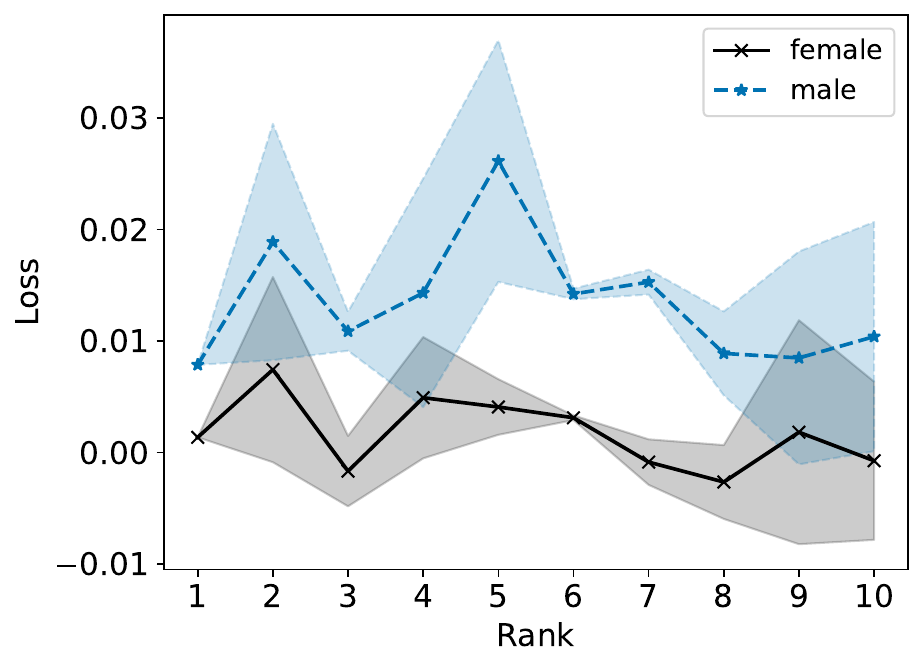}
\caption{Standard NMF applied on the entire Heart Disease data matrix. The reconstruction loss of each group is reported for ranks 1 to 10.}
\label{fig:snmf_hd_loss}
\end{figure}

\begin{figure}[h!]
\centering
\includegraphics[height=2.2in]{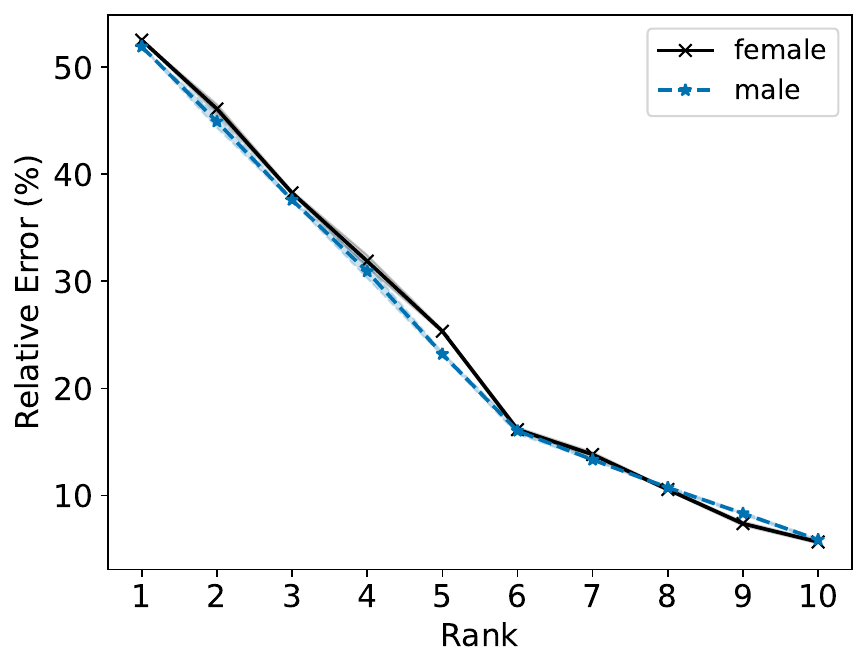} \includegraphics[height=2.2in]{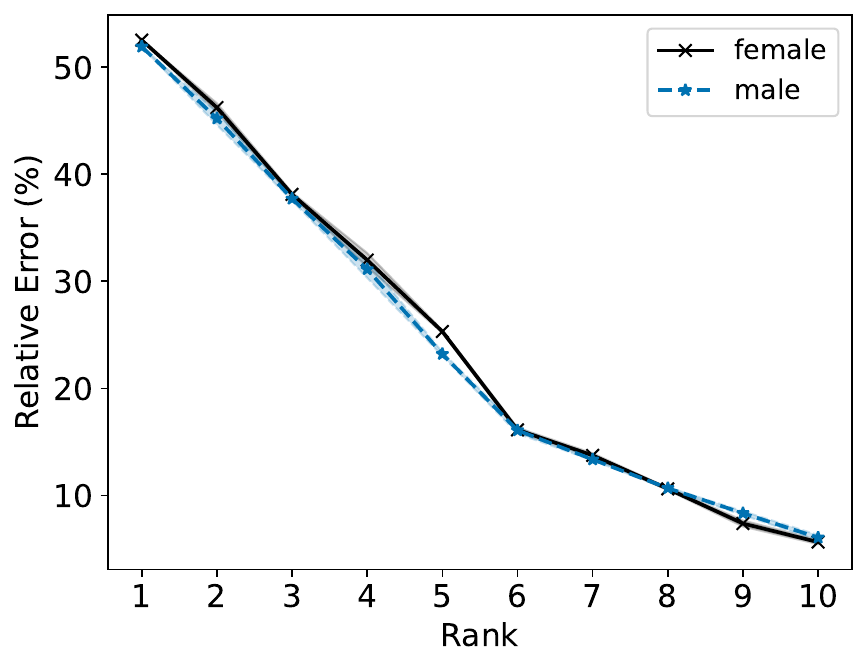}
\caption{\fnmf applied on the Heart Disease data matrix.
The Relative Error (\%) of each group is reported for ranks 1 to 10.
Left: \fnmf with the alternating minimization scheme).
Right: \fnmf with the multiplicative updates scheme.}
\label{fig:fnmf_hd_error}
\end{figure}
\begin{figure}[h!]
\centering
\includegraphics[height=2.2in]{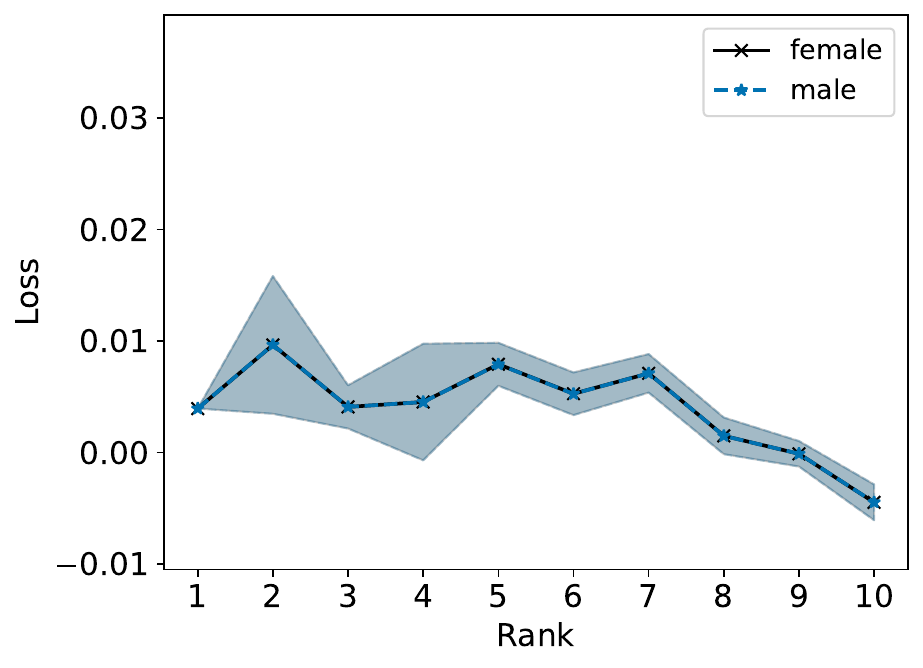} \includegraphics[height=2.2in]{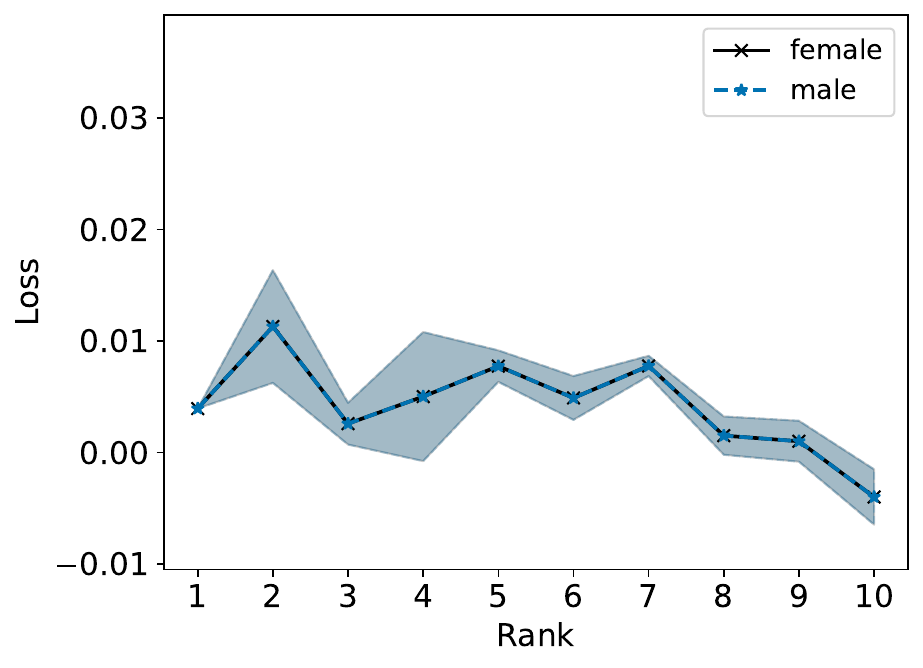}
\caption{\fnmf applied on the Heart Disease data matrix.
The reconstruction loss of each group in the dataset is reported for ranks 1 to 10.
Left: \fnmf with the alternating minimization scheme.
Right: \fnmf with the multiplicative updates scheme.}
\label{fig:fnmf_hd_loss}
\end{figure}

We seek to investigate the performance of standard NMF in representing the population stratified by patient sex (reported only as male or female in this dataset).
We conduct this analysis in an unsupervised setting, excluding the binary target variable that indicates disease presence or absence.
We also omit the sex attribute to perform our analysis on the two populations.
The numerical features in the dataset are non-negative and the categorical features are recorded as integers.  
There are $201$ individuals in the female group and $96$ in the male group.

In the right plot of \cref{fig:snmf_hd_error}, we observe that the male population generally incurs a slightly lower R-Error than the female population for the NMF models with ranks 1--5.
In the left plot of \cref{fig:snmf_hd_error}, NMF achieves similar R-Error for both groups for ranks 1--5 and lower R-Error for the female group compared to the male group.
In \cref{fig:snmf_hd_loss}, we observe overall higher R-Loss for the male group compared to the female group which indicates that NMF inadvertently favored the female group.

In \cref{fig:fnmf_hd_error} (left and right), we observe the \fnmf R-Error values are similar to the R-Error values when NMF is applied to each group individually (right plot of \cref{fig:snmf_hd_error}).
As observed in \cref{fig:fnmf_hd_loss}, overall \fnmf (MU and AM) achieves a similar loss for both populations which in this application may or may not be ``fair''.
With the fairness criterion considered in \fnmf, some patients will incur a higher reconstruction error compared to that achieved with a standard NMF model (e.g., for ranks~1--5).

We highlight that a possibility discussed in \cref{sec:estimating_E-ell} occurs for this dataset: it is possible for \fnmf to produce reconstructions with negative loss.  We see in \cref{fig:snmf_hd_loss,fig:fnmf_hd_loss} that this does happen at some of the ranks for at least one of the groups.  Indeed, in \cref{fig:fnmf_hd_loss} we see that the loss for a rank-$10$ \fnmf is consistently below $0$.  This shows that sometimes in practice \fnmf is able to find a better reconstruction for each group when applied to the full dataset than standard NMF can even when applied to just a single group.  In this case, the ``cost'' of representing both groups together as opposed to individually is negligible when using \fnmf.

\subsection{20Newsgroups Dataset}

The 20newsgroups dataset~\citep{20newsgroups} is a popular benchmark dataset containing documents gathered from $20$ newsgroups that are partitioned into $6$ major subjects. 
We sample $1500$ documents from the entire dataset with the number of samples from each subject is proportional to the size of the subject in the entire dataset. 
We cast all letters to lowercase and remove special characters as part of the pre-processing of the data.
The TFIDF vectorizer with the English stop words list from the NLTK package is applied to transform the text data into a matrix. 
After obtaining the data matrix of the entire dataset, the matrix is partitioned into $6$ groups according to the $6$ subjects present in the original dataset.
The sizes of the groups are: Computer $389$, Sale $78$, Recreation $316$, Politics $209$, Religion $193$, Scientific $315$.

\begin{figure}[h!]
\centering
\includegraphics[height=2.2in]{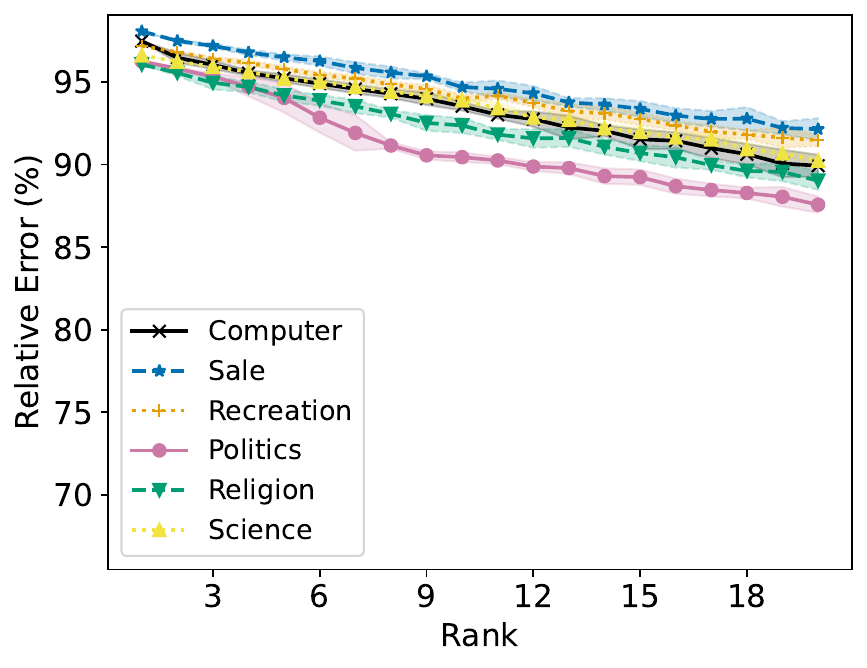} \includegraphics[height=2.2in]{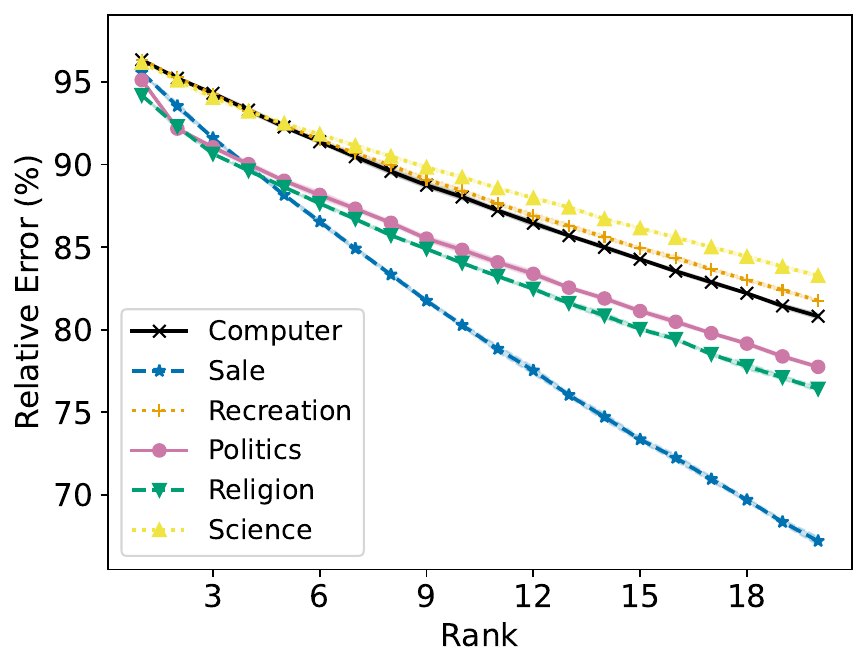}
\caption{
The Relative Error (\%) of each group in the 20 Newsgroups dataset is reported for ranks 1 to 20.
Left: standard NMF applied on the entire 20 Newsgroups data matrix. 
Right: standard NMF applied on the 20 Newsgroups data matrix of each group individually.}
\label{fig:snmf_20n_error}
\end{figure}

\begin{figure}[h!]
\centering
\includegraphics[height=2.2in]{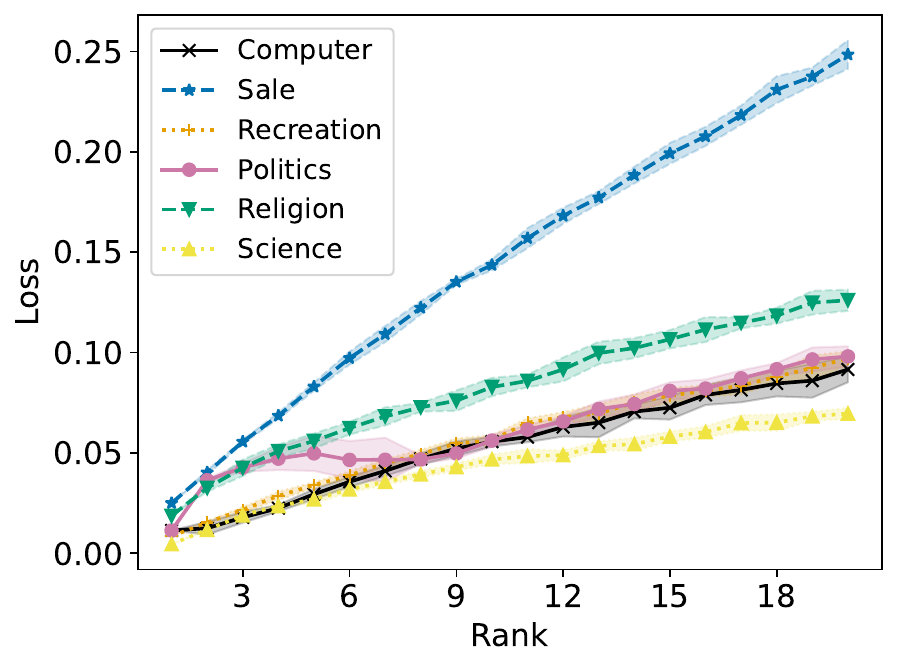}
\caption{Standard NMF applied on the entire 20 Newsgroups data matrix. The reconstruction loss of each group is reported for ranks 1 to 20.}
\label{fig:snmf_20n_loss}
\end{figure}

\begin{figure}[h!]
\centering
\includegraphics[height=2.2in]{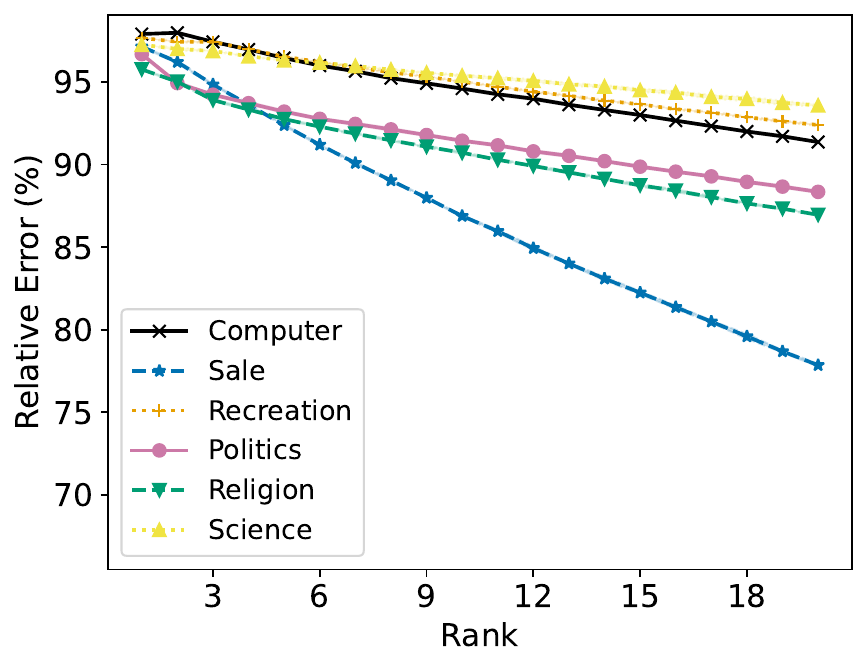} \includegraphics[height=2.2in]{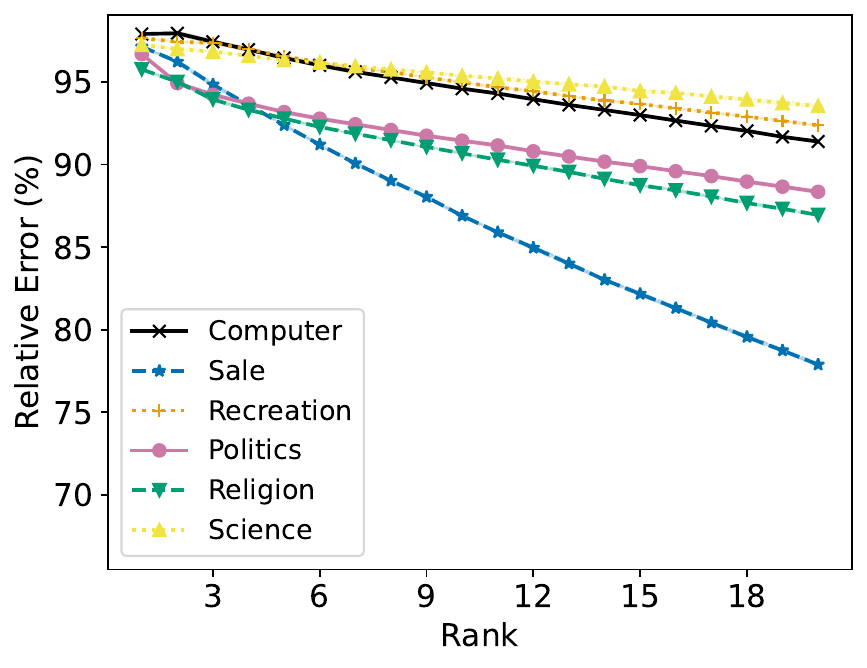}
\caption{\fnmf applied on the 20 Newsgroups data matrix.
The reconstruction loss of each group in the dataset is reported for ranks 1 to 20.
Left: \fnmf with the alternating minimization scheme.
Right: \fnmf with the multiplicative updates scheme.}
\label{fig:fnmf_20n_error}
\end{figure}

\begin{figure}[h!]
\centering
\includegraphics[height=2.2in]{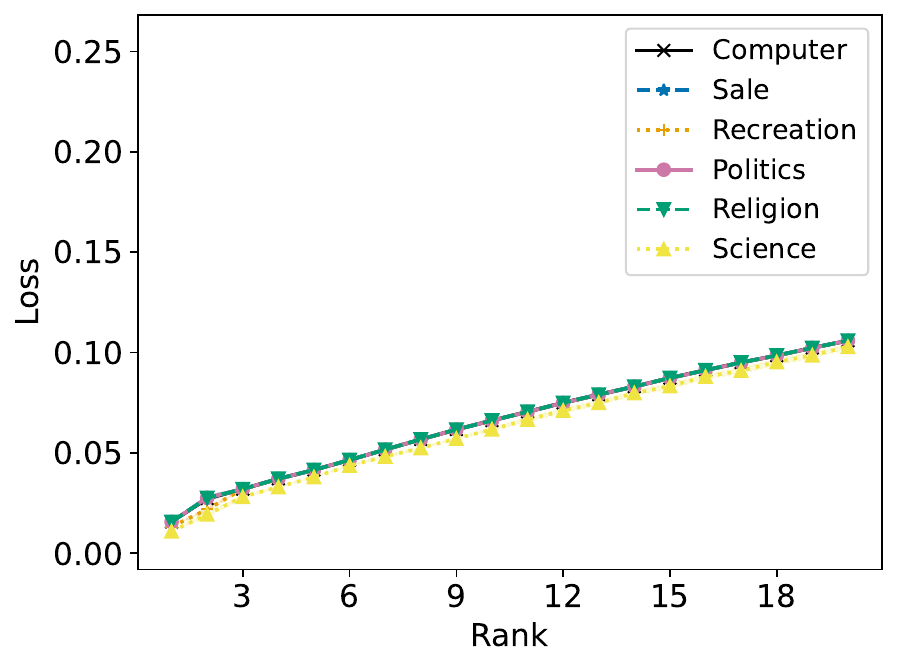} \includegraphics[height=2.2in]{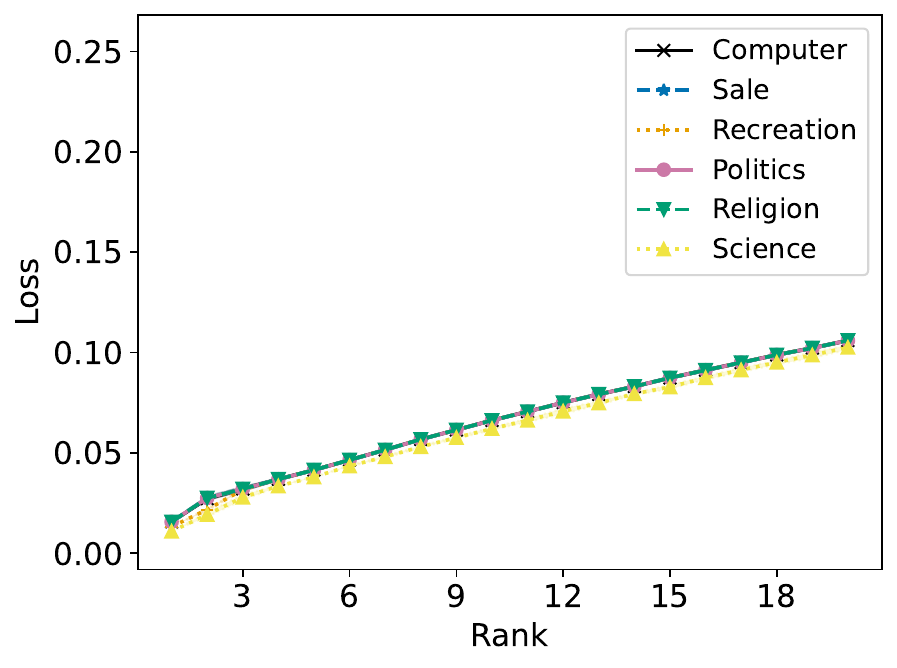}
\caption{\fnmf applied on the 20 Newsgroups data matrix.
The reconstruction loss of each group in the dataset is reported for ranks 1 to 20.
Left: \fnmf with the alternating minimization scheme.
Right: \fnmf with the multiplicative updates scheme.}
\label{fig:fnmf_20n_loss}
\end{figure}

\Cref{fig:snmf_20n_error} shows the  R-Error of each group with NMF, both applied to the full dataset at once and each group individually.  When NMF is performed on each group individually, the R-Error at rank 20 ranges from around 70\% to 85\%, depending on the group.  When one NMF reconstruction is obtained for the full dataset, the range of R-Error is more tightly clustered around 90\%.  Notably, the ``Sale'' group has the lowest R-Error when an NMF model is trained on each group individually, but the highest when one is trained on all groups at once.  This results in it having the highest R-Loss, as shown in \cref{fig:snmf_20n_loss}.

\fnmf is able to resolve this discrepancy.  \Cref{fig:fnmf_20n_error} (left and right) show that the six different groups have reconstruction errors that correspond to the individual group reconstruction errors.  
The ``Sale'' group is the group with the lowest R-Error, and the rest of the groups appear in the order they do in the right plot of \cref{fig:snmf_20n_error}.  Accordingly, the loss of each group under \fnmf (MU and AM), as shown in \cref{fig:fnmf_20n_loss}, are all very similar.

\Cref{fig:fnmf_20n_error,fig:fnmf_20n_loss} show another counterintuitive phenomenon.  
While R-Error for each group decreases as the number of ranks in the decomposition increases, R-Loss increases as the rank goes up.  
This is because the reconstruction error of each group decreases in rank much faster when decomposed individually than when decomposed together.

\subsection{Algorithm Comparisons}
\label{sec:alg-comparisons}

We propose two different algorithms for \fnmf: the alternating minimization method (\cref{alg:AM-fairnmf}) and the multiplicative updates method (\cref{alg:MU-fairnmf}).  In \cref{fig:fnmf_ortho_1_loss,fig:fnmf_ortho_2_loss,fig:fnmf_hd_loss,fig:fnmf_20n_loss}, both algorithms are run on three different datasets: the first synthetic dataset, the second synthetic dataset, the heart disease dataset, and the 20Newsgroups dataset, respectively.  For the heart disease and 20Newsgroups datasets, both algorithms perform equally well.  However, for the synthetic dataset, the alternating minimization method is more consistent in finding low loss solutions for the optimization problem.  This discrepancy is mirrored in \cref{sec: fnmf-algorithms}, where we only show a non-increasing result for the alternating minimization method.

\begin{figure}[h!]
\centering
\includegraphics[height=2.2in]{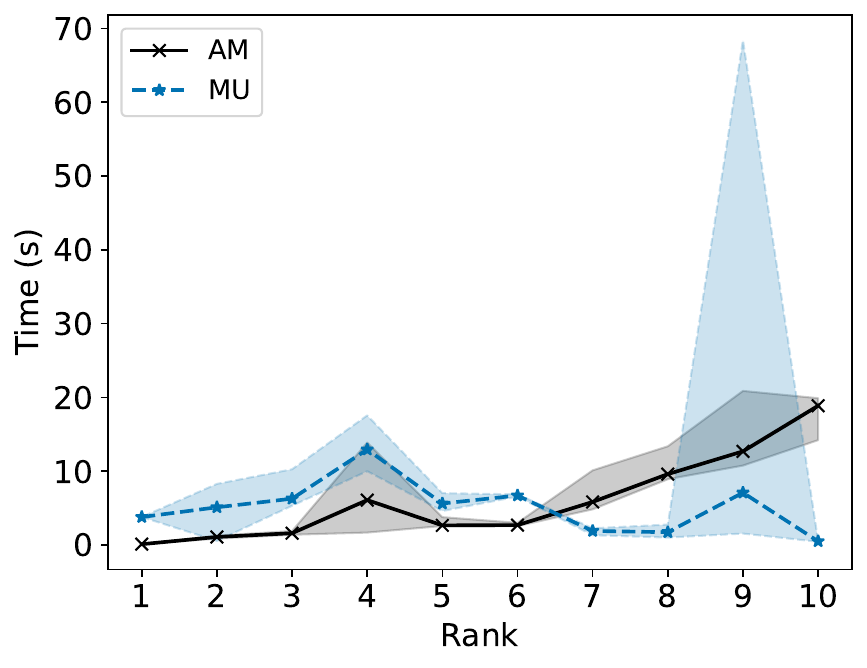}
\includegraphics[height=2.2in]{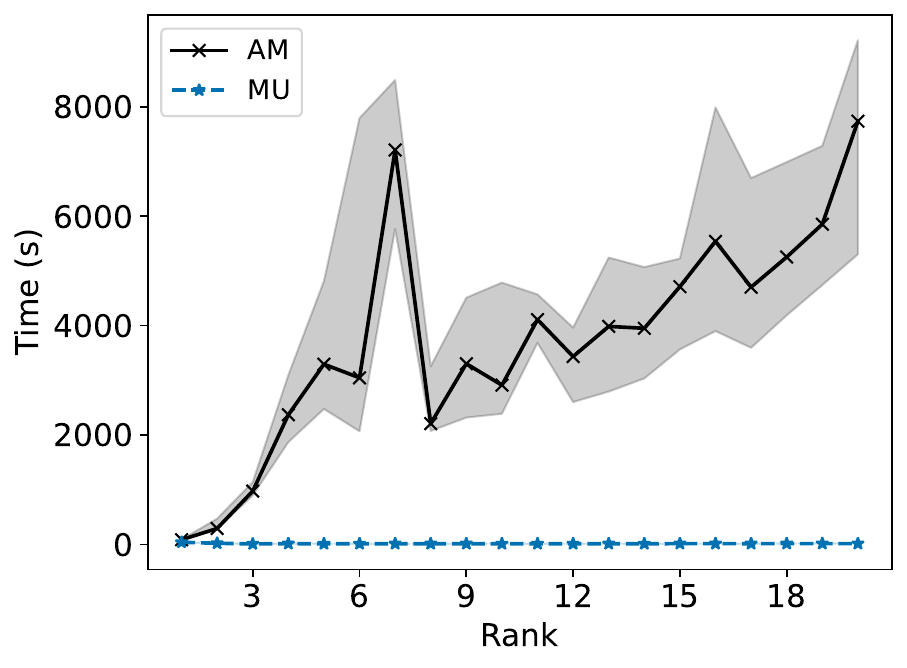}
\caption{Time in seconds required for \fnmf with alternating minimization (AM) scheme and with multiplicative updates (MU) scheme to converge for the Heart Diseases dataset (left) and 20Newsgroups dataset (right).  The convergence criterion is given by \cref{eq:convergence_condition}.  We report the median and interquartile range over 10 trials.}\label{fig:times}
\end{figure}

However, an important consideration when choosing between these two methods is the computation cost of each.  The alternating minimization involves solving one SOCP and one NNLS problem with each iteration, which is very expensive.  On the other hand, the multiplicative updates scheme only requires a few matrix multiplications in each iteration.  \Cref{fig:times} shows how long both algorithms took to reach convergence for the two real-world datasets\footnote{We do not report times for the synthetic datasets, due to the different choice in stopping condition.} when run on a 12 core 3.50GHz Intel i9-9920X CPU.  For the larger dataset (the 20Newsgroups dataset), the alternating minimization scheme is substantially slower than the multiplicative updates scheme.  A single \fnmf decomposition with the alternating minimization method can easily take over an hour to perform, whereas the longest time for convergence with the multiplicative updates method across all datasets and ranks is $129$ seconds.

To conclude this comparison, we refer back to \cref{fig:snmf_ortho_1_loss,fig:snmf_ortho_2_loss}.  Standard NMF applied to the second synthetic dataset sees the same variance as the MU rule for \fnmf, and on the first synthetic dataset the variance can also be high.  Furthermore, the MU rule is typically able to produce a lower mean loss for the highest loss group than standard NMF is typically capable of.  Even though MU rule is not as consistent in finding low loss solutions as the AM algorithm, it still compares favorably to standard NMF on all datasets.  As the MU algorithm is significantly faster than the AM algorithm, we expect it to be the preferred choice in most instances.  The AM algorithm should be used for critical applications, for small datasets, or when computation cost is not a concern.  In the case where high-quality reconstructions are critical but computation cost is a concern, we note that the only source of randomness in the MU algorithm is the initialization.   The MU rule can be modified by adjusting the update rule for $\bm c$ to decay more slowly or by starting at a different initialization for the matrices.


\section{Discussion}\label{sec:discuss}

We remark here on the title of the manuscript, and the notion that our proposed framework, and indeed any machine learning framework, is very unlikely to ever be completely ``fair''.  
On the other hand, there is certainly a need to make ML algorithms \textit{fairer} to help practitioners identify inequities and provide alternative methods that offer a fairer outcome for some applications.
Our objective in~\cref{eq:fnmf}, for example, asks that the maximum reconstruction loss across all population groups be minimized. 
In many contexts, as motivated in \cref{sec:fairer-NMF}, this results in fairer outcomes. 
Indeed, in many settings, populations consist of majority groups and minority groups, and because typical models minimize average or overall error, minority groups will typically have a higher reconstruction error than the majority. 
Further, standard NMF does not take into account the complexity of the groups.

We also comment here on the assumption that the $L$ population subgroups are known a priori. In many settings, this will be the case; for example, when the groups are defined by a specific population demographic within the dataset itself, this is a reasonable way of assigning subgroups.  In other settings, the groups may be more complex, or such information may simply be unavailable. We briefly remark on two possible avenues to address these settings, emphasizing again that we view the contribution of this manuscript as the proposal of and derivation of a fairer NMF formulation and not the identification of relevant subgroups (which we believe itself could warrant an entire study).

One possible way to learn the subgroups is by using NMF or another clustering method itself. Indeed, NMF is by its nature designed to learn topics that separate variables and/or the population. One could employ simple thresholding of the learned topic values (i.e. the values in $\mat W$), or apply a clustering method to the data or topic representations to divide the population according to topic strengths. It would of course be of interest to investigate whether such pre-processing into learned subgroups improves fairness when the groups are known a priori and can be compared.  A second alternative would be to apply our proposed fairer NMF method iteratively; that is, run standard NMF and divide the population into groups according to reconstruction error values (a group then consists of data points sharing similar error values).  Then those groups can be used, and this process could even be iteratively refined.  Of course, when there are no known groups, the notion of fairness itself changes and is highly application dependent. This is indeed why there are many notions of fairness to begin with, including individual fairness, which would align with extremely fine grained group structure.

The objective we propose takes into account the size and complexity of the groups and achieves fairer outcomes in these settings.
However, some drawbacks need to be considered, as will be the case for any method that attempts to mitigate fairness.  
First, we assume the population groups are known a priori. 
This can likely be overcome by learning the groups on the fly through cross-validation of reconstruction errors and is a future direction of research.  
The next concern of course is that it may not always be desirable to minimize the maximum reconstruction loss.  
Indeed, through this fairness mitigation, some groups and therefore individuals may receive a higher reconstruction error than they would have without the ``fairer'' approach. 
In settings like medical applications, where these tools are used to predict, for example, the likelihood of a patient having a disease, this may no longer seem fairer.  
It is thus clear that the notion of fairness itself is highly application-dependent, and great care should be taken when mitigating---or not---in learning methods.
A valuable direction for future work is to propose alternative NMF formulations under different fairness criteria and study their effectiveness across various settings and applications. Additionally, comparative evaluation of different fair GLRM frameworks (e.g., \citet{buet2022towards}) would be valuable, particularly in understanding the trade-offs between different fairness criteria and optimization approaches.

\section{Conclusion} 
\label{sec:Conclusion}
NMF is a widely used topic modeling technique in various domains, particularly when interpretability and trust are essential. 
We believe that examining the fairness of NMF is a valuable contribution to the field and an important step toward tackling key issues related to bias and fairness.
In this work, we presented an alternative NMF objective that seeks a non-negative low-rank model that provides equitable reconstruction loss across different groups. 
The goal is to learn a common NMF model for all groups under the min-max fairness framework which seeks to minimize the maximum of the average reconstruction loss across groups.
We proposed an alternating minimization algorithm and a multiplicative updates algorithm. 
Numerically, the latter demonstrated reduced computational time compared to a CVXPY~\citep{cvxpy} implementation of the AM algorithm while still achieving similar performance.
We showcased on synthetic and real datasets how standard NMF could lead to biased outcomes and discussed the overall performance of \fnmf.

\iftoggle{frontierspaper}{
\section*{Funding}}
{\section*{Acknowledgements}}
EG is partially supported by the UCLA Racial Justice seed grant, UCLA Dissertation Year Award, and the NSF Graduate Research Fellowship under grant DGE 2034835. LK and DN are partially supported by the Dunn Family Endowed Chair fund.  All authors were partially supported by NSF DMS 2408912.  This material is based upon work supported by the National Science Foundation under Grant No.\ DMS-1928930
and by the Alfred P.\ Sloan Foundation under grant G-2021-16778, while the authors EG and DN were in residence
at the Simons Laufer Mathematical Sciences Institute (formerly MSRI) in Berkeley, California, during the Fall 2023 semester.
\iftoggle{frontierspaper}{
\section*{Contribution}
LK contributed to the experiments and to the writing of the paper.  EG contributed to the experiments and designed the algorithms used.  DN supervised the work.  HG, NJN, and AL all assisted with the experiments.
}{}

\bibliography{bib}
\iftoggle{frontierspaper}{
\bibliographystyle{Frontiers-Harvard}
}{
\bibliographystyle{tmlr}
}

\end{document}